\documentclass[journal]{IEEEtran}
\usepackage{ifpdf}

\usepackage{cite}

\ifCLASSINFOpdf
  \usepackage[pdftex]{graphicx}
  \graphicspath{{imgfiles/}}
\else
\fi
\usepackage{algorithmic}

\usepackage[section]{placeins}
\newcommand{\ie}{\textit{i.e., }}
\newcommand{\eg}{\textit{e.g., }}
\newcommand{\etal}{\textit{et al.}}
\usepackage[dvipsnames, table]{xcolor}

\usepackage[cmex10]{amsmath}

\usepackage{amssymb}     
\usepackage{amsfonts}    
\usepackage[linesnumbered,ruled,vlined]{algorithm2e}

\usepackage{slashbox}
\usepackage[export]{adjustbox}
\usepackage{hyperref}
\usepackage{graphicx}     

\usepackage{subcaption}   
\usepackage{xcolor}       
\usepackage{multirow}     
\usepackage{hhline}       
\usepackage{makecell}     

\usepackage{array}

\usepackage{stfloats}
\begin{document}
%
\title{On the Adversarial Vulnerabilities of Transfer Learning in Remote Sensing}
%
%
%

\author{Tao~Bai*, Xingjian Tian*, Yonghao Xu, and~Bihan~Wen,~\IEEEmembership{Senior Member,~IEEE}
\thanks{T. Bai, X. Tian and B. Wen are with the School of Electrical and Electronic Engineering, Nanyang Technological University, Singapore 639798 (Email: tao.bai@ntu.edu.sg; xingjian.tian@outlook.com; bihan.wen@ntu.edu.sg).}
\thanks{Y. Xu is with the Computer Vision Laboratory (CVL) at the Department of Electrical Engineering (ISY), Linköping University, Linköping, Sweden. (Email: yonghao.xu@liu.se).}
\thanks{* Equal contribution.}
\thanks{This work has been submitted to the IEEE for possible publication. Copyright may be transferred without notice, after which this version may no longer be accessible.}}

%
%


\markboth{SUBMITTED TO IEEE TRANSACTIONS ON GEOSCIENCE AND REMOTE SENSING}{txj}
%



\maketitle

\begin{abstract}

The use of pretrained models from general computer vision tasks is widespread in remote sensing, significantly reducing training costs and improving performance. However, this practice also introduces vulnerabilities to downstream tasks, where publicly available pretrained models can be used as a proxy to compromise downstream models.
This paper presents a novel Adversarial Neuron Manipulation method, which generates transferable perturbations by selectively manipulating single or multiple neurons in pretrained models. Unlike existing attacks, this method eliminates the need for domain-specific information, making it more broadly applicable and efficient.
By targeting multiple fragile neurons, the perturbations achieve superior attack performance, revealing critical vulnerabilities in deep learning models. 
Experiments on diverse models and remote sensing datasets validate the effectiveness of the proposed method.
This low-access adversarial neuron manipulation technique highlights a significant security risk in transfer learning models, emphasizing the urgent need for more robust defenses in their design when addressing the safety-critical remote sensing tasks.
\end{abstract}

\begin{IEEEkeywords}
Adversarial Attack, Neuron Manipulation, Remote Sensing, Transfer learning, Image Classification, Adversarial Robustness
\end{IEEEkeywords}

%
\IEEEpeerreviewmaketitle

\section{Introduction}

\IEEEPARstart{D}{eep} learning has demonstrated exceptional performance across a wide range of tasks in the field of remote sensing and Earth observation, revolutionizing how we analyze and interpret spatial data. 
Tasks such as land-use classification~\cite{Shreehari2024land}, object detection~\cite{liu2022object}, semantic segmentation~\cite{kaiser2017learning}, and change detection~\cite{lin2019multispectral} have benefited significantly from the powerful feature extraction capabilities of deep neural networks. However, despite its potential, the widespread adoption of deep learning in remote sensing faces two major challenges: data scarcity and the high computational costs associated with training models from scratch. High-quality labeled data for remote sensing applications are often limited due to the specialized expertise required for annotation, while training models on complex remote sensing data, such as high-dimensional hyperspectral imagery or very high-resolution (VHR) imagery with fine spatial details, demands significant computational resources.

\begin{figure}[!t]
    \centering
    \includegraphics[width=\linewidth]{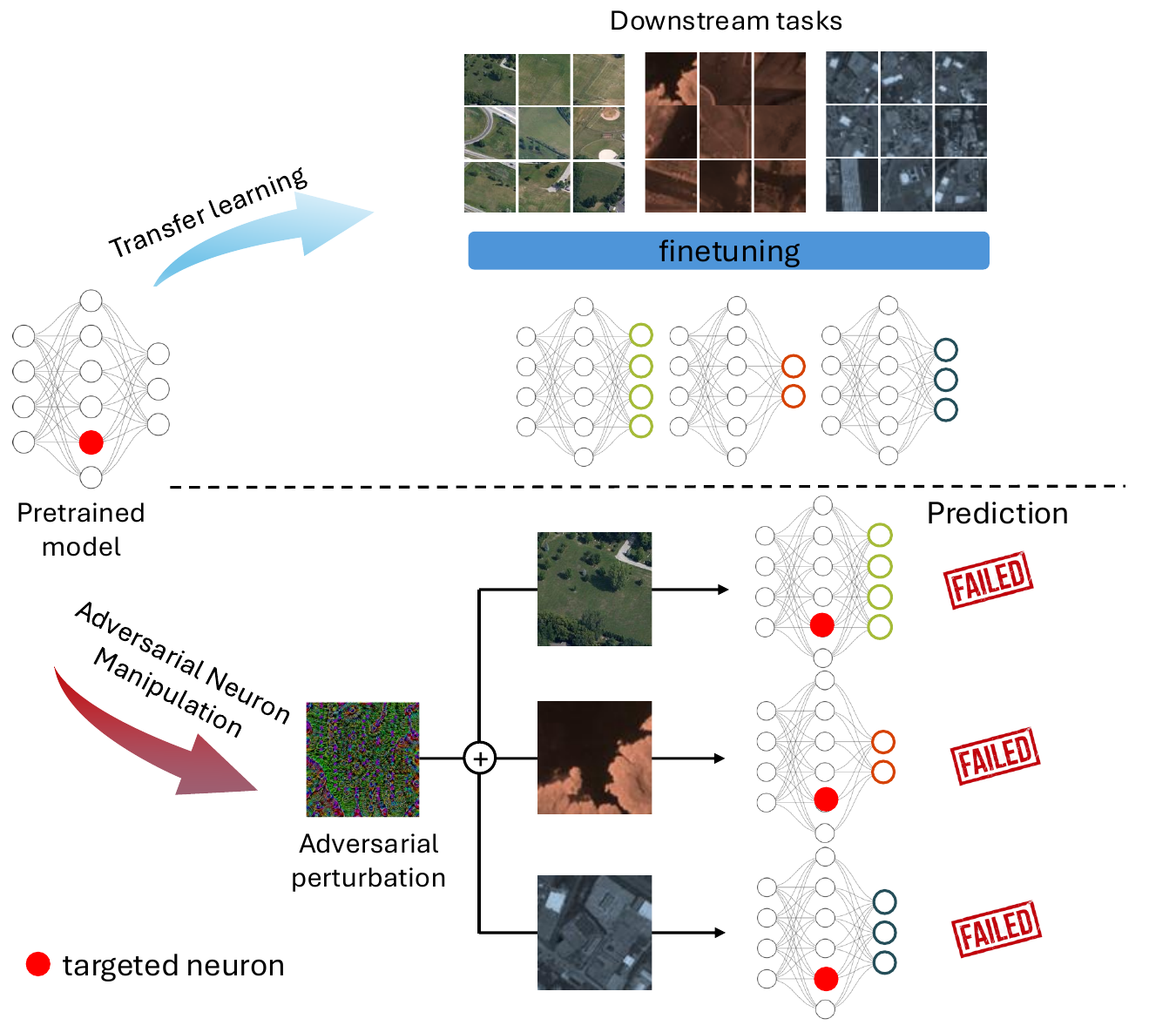}
\caption{The adversarial perturbation generated on publicly available pretrained deep models with Adversarial Neuron Manipulation(ANM) can threaten transfer-learned models effectively. 
}
    \label{fig1}
\end{figure}

To address these challenges, the use of pretrained models from other domains has become a prevalent strategy, a technique widely referred to as transfer learning~\cite{ma2024transfer,xie2016transfer,liu2020similarity}. By leveraging models pretrained on large, diverse datasets (e.g., ImageNet), researchers can fine-tune these models for specific remote sensing tasks, significantly reducing the need for extensive labeled data and computational power. This approach has enabled the remote sensing research community to achieve remarkable progress in various applications. For instance, transfer learning has been successfully employed for land-use classification with high spatial resolution (HSR) imagery~\cite{liu2020similarity}, building extraction~\cite{ait2023enhancing}, and change detection in VHR images~\cite{liu2021building}, among other tasks. These advancements underscore the transformative impact of transfer learning in overcoming barriers to the adoption of deep learning in remote sensing.


Deep learning models, nevertheless, have been consistently demonstrated their vulnerability to adversarial attacks, a challenge that has been well-documented in the literature~\cite{madry2017towards,goodfellow2014explaining,carlini2017towards}. Adversarial attacks involve the introduction of carefully crafted perturbations to input data, which are often imperceptible to human observers but can cause deep learning models to produce incorrect outputs with high confidence. This susceptibility raises serious concerns, particularly in critical applications where the reliability and robustness of deep learning models are paramount.
In response to these challenges, the remote sensing research community has begun to explore the implications of adversarial attacks within its domain~\cite{xu2023ai}. For example, Bai \etal~\cite{bai2022targeted} proposed a targeted adversarial attack tailored specifically for the scene classification task in remote sensing. Extending this work, Bai \etal~\cite{10471615} investigated adversarial attacks on semantic segmentation tasks, highlighting how these perturbations can disrupt the segmentation of remote sensing imagery. Furthermore, Xu and Ghamisi~\cite{xu2022universal} developed a novel approach for generating universal adversarial examples, which are capable of simultaneously compromising multiple tasks and models. These advancements underscore the pressing need to understand and mitigate adversarial vulnerabilities in remote sensing applications.
For a systematic and comprehensive review of adversarial attacks in remote sensing, readers are referred to~\cite{xu2023ai}, which consolidates the existing research and highlights key challenges and future directions.

Surprisingly, the adversarial risks associated with using pretrained models have been largely overlooked by the remote sensing community. Popular models such as VGG~\cite{simonyan2014very}, ResNet~\cite{he2016deep}, U-Net~\cite{DBLP:journals/corr/RonnebergerFB15}, EfficientNet~\cite{tan2019efficientnet}, and MobileNet~\cite{howard2019searching} are commonly employed as the foundation for remote sensing tasks. However, this widespread adoption creates a vulnerability that attackers can exploit. Specifically, adversaries can leverage publicly available pretrained models to craft adversarial attacks without requiring knowledge of the specific remote sensing applications where these models are deployed.

In this paper, we highlight above risks and introduce \textit{\textbf{A}dversarial \textbf{N}euron \textbf{M}anipulation}~\textbf{(ANM)}, a novel adversarial attack which generates effective adversarial examples by manipulating neurons within the model (see Figure~\ref{fig1}). 
Unlike existing adversarial attack methods~\cite{bai2022targeted,xu2022universal}, which typically require access to the target models and their associated data, ANM specifically targets publicly available pretrained models and operates independently of downstream tasks or training data. This independence makes ANM more practical and adaptable for real-world scenarios.
ANM specifically targets publicly available pretrained models, avoiding assumptions about downstream tasks and data. By activating individual or multiple neurons within these models, ANM produces strong adversarial perturbations that can compromise models fine-tuned from the original pretrained architectures. We demonstrate how these perturbations can effectively mount successful attacks against fine-tuned models in various downstream remote sensing tasks, thereby exposing the hidden vulnerabilities of pretrained models in this domain.

In summary, this paper makes the following key contributions:
\begin{itemize}
    \item 
    To the best of our knowledge, this is the first work to systematically investigate the adversarial risks associated with the adoption of pretrained models in remote sensing. We uncover these vulnerabilities and provide a detailed demonstration of how attackers can exploit them, emphasizing the critical need for increased awareness and robust defenses.
    \item 
    We propose Adversarial Neuron Manipulation (ANM), a groundbreaking attack technique that targets fine-tuned models derived from publicly available pretrained models. By selectively activating specific single neuron within the pretrained models, ANM generates strong adversarial perturbations while remaining agnostic to the downstream tasks, fine-tuned models, and associated training data, making it highly practical and adaptable.
    \item 
    We further propose ANM-M, which targets at a group of neurons, to enhance the transferable capabilities of our attack. Through extensive experiments across diverse models and datasets, ANM-M achieves consistently high accuracy drops of unseen models (48\% on UCM and 60\% on AID), showcasing its effectiveness and generalizability.
\end{itemize}





This paper is organized as follows. Section \ref{sec:RelatedWorks} introduces related works and discusses existing black-box attack methods on transfer learning models and their limitations. Section \ref{sec:Metho} elaborates on a common risk of transfer learning models, namely the grey-box threat scenario. It further presents the details of the proposed attack methods based on adversarial neuron manipulation. 
In Section \ref{sec:exp}, we describe the experimental setup, including the datasets, victim models, and implementation details of ANM-S and ANM-M. This section also evaluates the effectiveness of the proposed attacks under various scenarios, demonstrating their transferability and impact across different architectures and datasets. 
Finally, Section \ref{sec:conclusion} concludes the paper by summarizing the key findings and discussing the broader implications of the proposed attack strategies for the remote sensing community. We also outline future research directions in this section.

\section{Related Works}\label{sec:RelatedWorks}
\subsection{Transfer Learning}\label{subsec:Transfer}

Transfer learning \cite{tl} leverages the highly transferable features extracted by deep models pretrained on extensive datasets like ImageNet \cite{deng2009imagenet}, facilitating a faster training process with relatively smaller datasets while achieving performance comparable to models trained from scratch \cite{kornblith2019better}. Practitioners can utilize pretrained networks, such as ResNet \cite{he2016deep} and VGG \cite{simonyan2014very}, and adapt them to new classification tasks, such as remote sensing image (RSI) classification, by training a simple fully connected (FC) layer or linear classifier using the features from the penultimate layer. 

In the context of transfer learning for RSI classification, the retrain-classifier-only approach is a widely used method, where retraining only the final one or two layers often yields comparable performance to fully fine-tuned models \cite{hu2015transferring}. Feature extractors based on convolutional neural networks (CNNs) pretrained on the ImageNet dataset have proven to be effective for transfer learning in earth observation and RSI classification tasks \cite{marmanis2015deep}. A single feature extractor can be utilized to train multiple downstream classifiers for different classification tasks using the extracted features. For instance, by keeping all layers of ResNet50 before the penultimate layer frozen as a feature extractor, it is possible to train separate downstream classifiers for distinct remote sensing classification tasks, such as crop classification \cite{munipalle2023agricultural} and building classification \cite{9696140}. This approach requires only two classifiers rather than training two complete models, thus improving efficiency. 

Pretrained models demonstrate strong generalization capabilities on remote sensing datasets, extracting meaningful feature maps that are valuable for image classification. Using a pretrained feature extractor can significantly reduce the number of trainable parameters, resulting in superior performance across multiple remote sensing datasets \cite{nogueira2017towards}. Castelluccio \etal \cite{castelluccio2015land} explored the application of convolutional neural networks (CNNs) pretrained on ImageNet for land-use classification in remote sensing. Their experiments demonstrated that adjusting solely the last layer(s) outperformed training the entire model from scratch, achieving high accuracy with substantially fewer training epochs.
Similarly, Hu \etal \cite{hu2015transferring} leveraged transfer learning for high-resolution RSI classification by re-training the final layer of a pretrained CNN. 
Tasks such as environmental remote sensing~\cite{ma2024transfer}, poverty mapping~\cite{xie2016transfer}, building extraction~\cite{ait2023enhancing}, building change detection~\cite{liu2021building}, and RSI semantic segmentation~\cite{cui2020semantic} can be effectively performed using transfer learning.

Additionally, transfer learning methods can be enhanced through approaches such as Active Transfer Learning, which incorporates active learning strategies to improve model performance \cite{deng2019active}, and multisource domain transfer learning, which effectively supports hyperspectral image classification across varied data sources \cite{yang2022multi}. Moreover, Geng \etal \cite{geng2020transfer} made significant advancements in applying transfer learning to synthetic aperture radar (SAR) remote sensing, demonstrating enhanced classification accuracy and model robustness through specialized transfer learning techniques tailored to SAR imagery.

\subsection{Adversarial Attacks}\label{subsec:White}

Adversarial attacks involve the use of carefully crafted images, known as adversarial examples, designed to induce abnormal responses in models, thereby reducing model accuracy, diminishing the confidence in certain classes, or causing incorrect predictions with high confidence \cite{goodfellow2014explaining}. Adversarial examples can significantly undermine the performance of deep learning models \cite{szegedy2013intriguing}, and their impact on machine learning systems has attracted increasing attention. Prominent adversarial attack methods, such as the Fast Gradient Sign Method (FGSM) \cite{goodfellow2014explaining} and Projected Gradient Descent (PGD) \cite{madry2017towards}, use gradient ascent on a given clean image through backpropagation to generate adversarial examples, leading the target model to produce incorrect predictions. 
An adversarial perturbation is the difference between an original image and its adversarial example, represented as a matrix with values close to zero to minimize visual differences.
Classic FGSM and PGD attack are white-box attacks that necessitate that the adversary possesses detailed knowledge of the network architecture and the trained weights to effectively generate adversarial perturbations. However, in practical scenarios, real-world deep learning models typically do not disclose all details to the public. Consequently, these white-box attacks may not be feasible in many situations. To address this limitation, several black-box attack methodologies have emerged in recent years.

Black-box attacks represent any attack strategies that do not fall within the white-box category \cite{mahmood2021back}. Generally, it refers to scenarios without knowledge of the model's internal mechanisms or the training data required \cite{papernot2017practical}. 
Query-based black-box attacks such as \cite{vo2022query,narodytska2017simple} assume that the attacker can observe the raw outputs or the probabilities vectors of the victim network.
Surrogate model-based black-box attacks such as local substitute model attack \cite{rahmati2020geoda} operate under the assumption that a subset of data from the training or validation set can be obtained, along with access to the prediction outputs from the target classifier.

Adversarial attacks pose a significant threat to geoscience and remote sensing tasks, where safety considerations are paramount \cite{xu2023ai}. Adversarial examples generated for satellite images have demonstrated substantial efficacy in compromising the performance of deep learning classification models \cite{czaja2018adversarial}. Xu~\etal \cite{9119167} found that adversarial examples can easily fool state-of-the-art deep neural networks designed for high-resolution RSIs. Chen \etal \cite{chen2021empirical} found that traditional white-box attack methods are effective in launching adversarial attacks against deep models for both optical and SAR RSIs. Bai~\etal~\cite{bai2022targeted} proposed strong targeted attacks using universal adversarial examples in the domain of  RSIs. UAE-RS \cite{xu2022universal} provides black-box adversarial
samples in the remote sensing field.
Czaja \etal \cite{czaja2018adversarial} introduced an approach to simulate physical attacks within digital space, using metadata to align the attack with remotely sensed data, which is unique in its attention to geospatial context. The Stealthy Attack for Semantic Segmentation (SASS) and its variant, Masked SASS (MSASS)~\cite{10471615}, achieve a high level of stealthiness and effectively reduce the detection success rate in segmentation tasks.
Refinements on classic attack methods are highly effective in compromising remote sensing applications. For example, the Scale-Adaptive Adversarial Patch method \cite{lu2021scale} generates adversarial examples for aircraft detection by addressing the challenge of significant scale variations in RSIs. Additionally, Zhang \etal \cite{zhang2022adversarial} explored physical adversarial attacks targeting multi-scale objects and object detection within unmanned aerial vehicle (UAV) remote sensing imagery.

\section{Methodologies}\label{sec:Metho}

\subsection{Problem Formulation}\label{subsec:preli}
In contrast to existing attack methodologies that typically assume full access to the target models and their associated data, this work introduces a novel \emph{grey-box threat scenario}. This scenario is defined by its reliance solely on publicly available pretrained models, without requiring supplementary information such as downstream classification details or training specifics. 
Within this framework, the constraints imposed by the inability to train synthetic models or simulate task-specific behaviors, as emphasized in \cite{rahmati2020geoda}, highlight the practical challenges of the approach. By focusing on this constrained yet realistic threat model, this study uncovers a fundamental vulnerability in transfer learning workflows.

Under the grey-box threat scenario, the attack begins with a public pretrained model, denoted as $f_p$. The primary objective is to generate adversarial perturbations, represented as $\delta$, that maximize the neuron activations of $f_p$. These perturbations are subsequently applied to attack a victim model, $f_d$, which is derived from $f_p$ by inheriting some of its pretrained weights. Importantly, while the perturbations $\delta$ are crafted using $f_p$, they are designed to target $f_d$, a distinct model. 
This approach is distinguished by its independence from knowledge about how $f_d$ is obtained from $f_p$. Unlike existing attacks, such as those described in \cite{bai2022targeted,rahmati2020geoda}, the proposed method does not require access to the specifics of the victim model’s fine-tuning or adaptation process, thereby setting it apart in its applicability and practicality.

\begin{figure*}[t]
    \begin{minipage}[t]{0.45\textwidth}
        \centering
        \includegraphics[width=\linewidth]{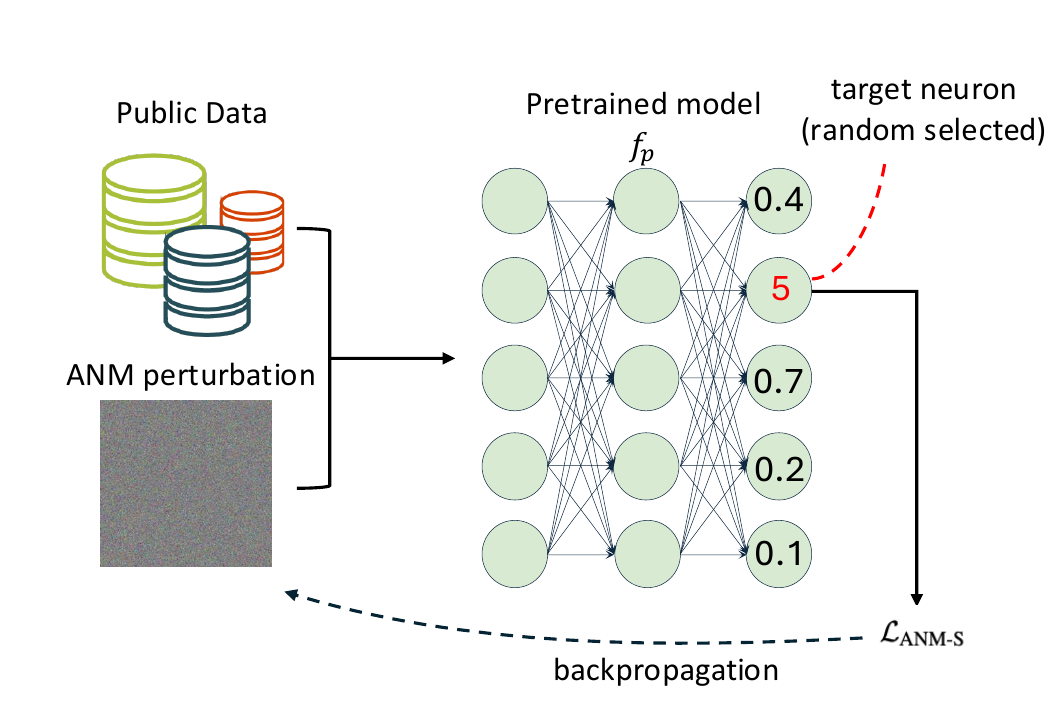}
        \captionof{figure}{Overview of ANM-S. The ANM perturbation is optimized to maximize the value of a single neuron, which is randomly selected.}
        \label{fig:anm-s}
    \end{minipage}%
    \hfill
    \begin{minipage}[t]{0.45\textwidth}
        \centering
        \includegraphics[width=\linewidth]{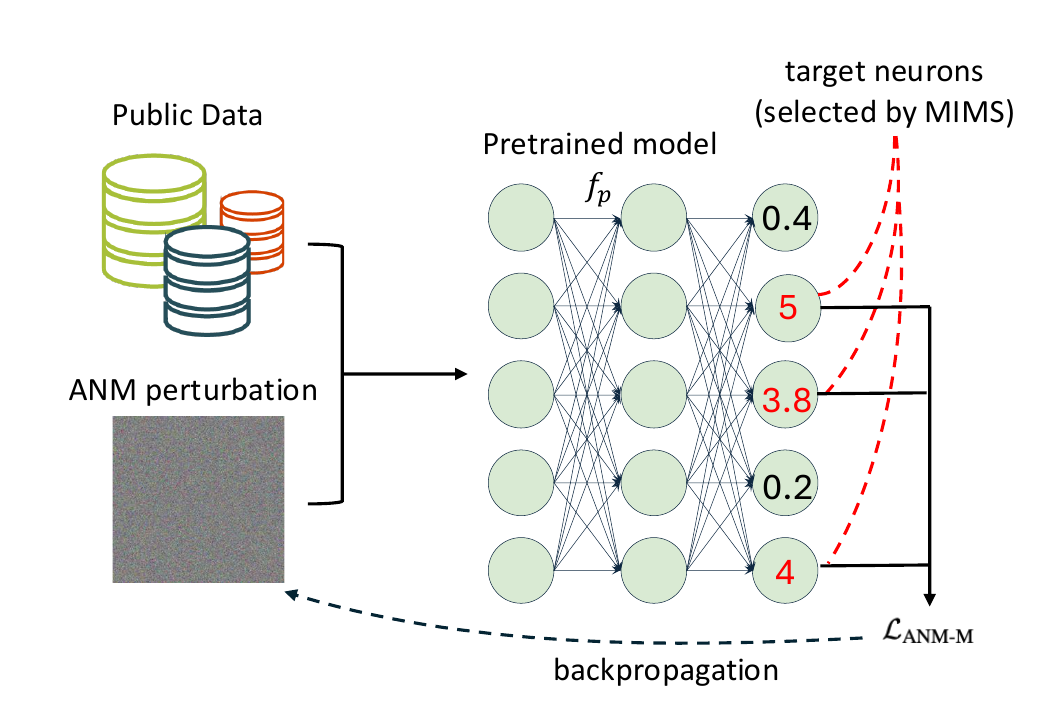}
        \captionof{figure}{Overview of ANM-M. The ANM perturbation is optimized to maximize the value of a group of neurons, which have the maximal inter-neuron MI and are obtained via Mutual Information Maximization Search (MIMS).}
        \label{fig:anm-m}
    \end{minipage}
\end{figure*}



\subsection{Adversarial Neuron Manipulation for Single Neuron (ANM-S)} \label{subsec:ptb1}

Unlike traditional adversarial attacks that often rely on fixed target models, we narrow our focus to a single neuron in pretrained models, making our method distinct from existing approaches. The proposed \emph{Adversarial Neuron Manipulation for Single Neuron (ANM-S)} method exploits the vulnerabilities of transfer learning workflows by targeting specific neurons in a pretrained model \( f_p \). As illustrated in Fig.~\ref{fig:anm-s}, this targeted approach is particularly effective in grey-box scenarios where access to the victim model \( f_d \) is limited or unavailable.

Let \( \mathbf{x} \) denote an input sample from a publicly available dataset \( \mathcal{D} \), and let \( f_p \) represent a publicly available pretrained model. Within \( f_p \), a neuron \( \phi \) in a specific layer is randomly selected as the target. The activation of this neuron for input \( \mathbf{x} \) is denoted as \( \phi(\mathbf{x}; f_p) \). The objective of ANM-S is to craft an adversarial perturbation \( \delta \) that maximizes the activation of \( \phi \), while ensuring that the perturbation remains imperceptible to human observers. Formally, this can be expressed as:
\begin{equation}
    \delta = \arg\max_{\delta, \|\delta\|_\infty \leq \epsilon} \phi(\mathbf{x} + \delta; f_p),
\end{equation}

where \( \epsilon \) is the perturbation budget that constrains the maximum allowable change to the input \( \mathbf{x} \).

To accelerate the optimization process while maintaining a high attack success rate, the loss function for ANM-S is defined as:
\begin{equation}
    \mathcal{L}_{\text{ANM-S}} = \mathbb{E}_{\mathbf{x} \in \mathcal{D}} \left[ \left( t - \phi_k\left(\mathbf{x} + \delta; f_p \right)   \right)^2 \right],
\end{equation}
where \( t \) is a hyperparameter that controls the intensity of the attack. The value of $t$ will be elaborated in Section~\ref{sec:exp}.

The perturbation \( \delta \) is generated using gradient-based optimization techniques such as \emph{Projected Gradient Descent (PGD)}. During each iteration, the gradient of the neuron activation \( \phi \) with respect to the input \( \mathbf{x} \) is computed, and \( \delta \) is updated to maximize \( \mathcal{L}_{\text{ANM-S}} \) while satisfying the \( \ell_\infty \) norm constraint:
\begin{equation}
    \delta^{(t+1)} = \text{Proj}_{\|\delta\|_\infty \leq \epsilon} \left(\delta^{(t)} - \eta \cdot \nabla_{\delta} \mathcal{L}_{\text{ANM-S}}\right),
\end{equation}
where \( \eta \) is the step size, and \( \text{Proj} \) is the projection operator that ensures the perturbation satisfies the norm constraint.

Once the perturbation \( \delta \) is crafted using \( f_p \), it is applied to the input sample \( \mathbf{x} \) to produce the adversarial example \( \mathbf{x}^* = \mathbf{x} + \delta \). This adversarial example is then evaluated on the victim model \( f_d \), which inherits some of the pretrained weights from \( f_p \). The key insight behind ANM-S is that the adversarial effect, crafted by targeting a specific neuron in \( f_p \), transfers effectively to \( f_d \), even without direct access to \( f_d \). The details of ANM-S are provided in Algorithm~\ref{algo:anm-s}.

\begin{figure}[t]
    \centering
    \begin{subfigure}[t]{0.45\columnwidth}
        \centering
        \includegraphics[width=\linewidth]{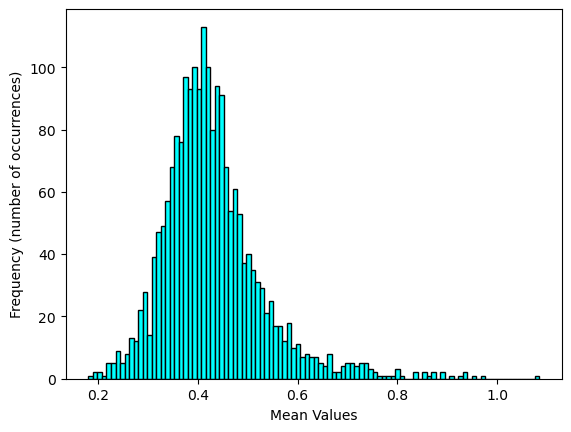}
        \caption{The histogram of mean values of the ResNet50 feature extractor.}
        \label{fig:mean}
    \end{subfigure}
    \hfill
    \begin{subfigure}[t]{0.45\columnwidth}
        \centering
        \includegraphics[width=\linewidth]{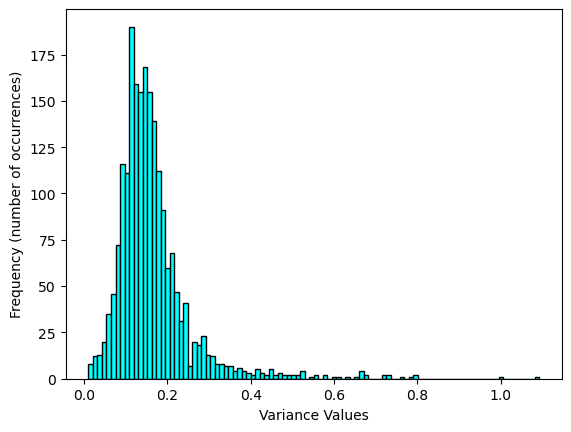}
        \caption{The histogram of variance of 2048 neurons in ResNet50 feature extractor.}
        \label{fig:var}
    \end{subfigure}
    \caption{Histograms of mean values and variances of ResNet50 feature extractor neurons.}
    \label{fig:norm_dist}
\end{figure}

\begin{algorithm}[t]
\caption{ANM-S}
\label{algo:anm-s}
\KwIn{
    Pretrained model \( f_p \), dataset \( \mathcal{D} \), perturbation budget \( \epsilon \), step size \( \eta \), maximum iterations \( T \), batch size \( B \), and hyperparameter \( t \).
}
\KwOut{
    Adversarial perturbation \( \delta \).
}
\BlankLine
\textbf{Initialization:} Randomly select a target neuron \( \phi \) from pretrained model \( f_p \). Set \( \delta^{(0)} \leftarrow 0 \)\;
\For{$i = 0$ \textbf{to} $T-1$}{
    Sample a mini-batch \( \mathcal{B} \subset \mathcal{D} \) of size \( B \)\;
    Compute the gradient of the loss function with respect to \( \delta \):
    \[
    \nabla_{\delta} \mathcal{L}_{\text{ANM-S}} = \frac{1}{|\mathcal{B}|} 
    \sum_{\mathbf{x} \in \mathcal{B}} \nabla_{\delta} \bigg[ 
    \big( t - \phi(\mathbf{x} + \delta^{(i)}; f_p)) ^2
    \bigg]
    \]
    Update the perturbation using Projected Gradient Descent (PGD):
    \[
    \delta^{(i+1)} = \text{Proj}_{\|\delta\|_\infty \leq \epsilon} \left(\delta^{(i)} - \eta \cdot \nabla_{\delta} \mathcal{L}_{\text{ANM-S}}\right)
    \]
}
\Return \( \delta^{(T)} \)\;
\end{algorithm}

\begin{algorithm}[t]
\caption{ANM-M}
\label{algo:anm-m}
\KwIn{
    A trained model \( f_p \), 
    dataset \( \mathcal{D} \), 
    an initial seed neuron \( \phi_0 \), 
    a budget \( K \) (maximum number of neurons to select), 
    and a perturbation constraint \( \epsilon \).
}
\KwOut{
    Selected neuron subset \( \Omega \) and adversarial perturbation \( \delta \).
}
\BlankLine
\textbf{Initialization:} Set \( \Omega \leftarrow \{\phi_0\} \)\;
Mutual Information Maximization Search (MIMS):\\
\While{\(|\Omega| < K\)}{
    \For{each candidate neuron \( \phi \notin \Omega \)}{
        Compute \( I(\Omega; \phi) = \frac{1}{2}\ln\left(\frac{\det(\Sigma_{\Omega})\det(\Sigma_{\phi})}{\det(\Sigma_{\Omega \cup \{\phi\}})}\right) \) using covariance matrices\;
    }
    Select \( \phi^* = \arg\max_{\phi \notin \Omega} I(\Omega; \phi) \)\;
    Update \( \Omega \leftarrow \Omega \cup \{\phi^*\} \)\;
}
\BlankLine
\For{$i = 0$ \textbf{to} $T-1$}{
    Sample a mini-batch \( \mathcal{B} \subset \mathcal{D} \) of size \( B \)\;
    Compute the gradient of the loss function with respect to \( \delta \):
    \begin{equation*}
    \begin{aligned}
 \nabla_{\delta} &\mathcal{L}_{\text{ANM-M}} = \\& \frac{1}{|\mathcal{B}|} 
\sum_{\mathbf{x} \in \mathcal{B}} \nabla_{\delta} \Biggr[ \frac{1}{K}\sum_{j=0}^{K-1}\bigg[ 
\big( t_j - \phi_j(\mathbf{x} + \delta^{(i)}; f_p)\big) ^2
\bigg]\Biggr]
\end{aligned}
\end{equation*}
    Update the perturbation using Projected Gradient Descent (PGD):
    \[
    \delta^{(i+1)} = \text{Proj}_{\|\delta\|_\infty \leq \epsilon} \left(\delta^{(i)} - \eta \cdot \nabla_{\delta} \mathcal{L}_{\text{ANM-M}}\right)
    \]
}
\Return \( \delta^{(T)} \)\;
\end{algorithm}

\begin{table}[t]
\caption{ Impact of Target Value Selection on ANM-S Attack Performance. 
The target is $2015^\text{th}$ neuron  of ResNet50. lr, $N_{drop}$ and $N$ represent the initial step size, number of epochs for step drop, and the total training epochs, respectively. Target Value is calculated by $\mu+k\sigma$ or set directly.}
\label{table:target_value}
\centering
\begin{tabular}{cccccc}
\noalign{\hrule height1pt}
 lr & \begin{math} N_{drop} \end{math}& \begin{math} N \end{math}  & Target Value  & ACC (\%) \\
\noalign{\hrule height.5pt}
 4$\epsilon$ & 2 & 10   &  0.2471(k=0)        & 86.48 \\
4$\epsilon$  & 2 & 10   & 0.4423(k=1)        &  58.19 \\
 4$\epsilon$  & 2 & 10 & 0.6375(k=5)        & 27.81  \\
 4$\epsilon$  & 2 & 10  & 2.1991(k=10)        & \ 5.33 \\
 8$\epsilon$  & 4 & 28 &  10  &17.62 \\
 8$\epsilon$  & 4  & 28  & 50 &  10.67 \\
16$\epsilon$  & 5 & 45 & 100  &  \ 8.76 \\
 16$\epsilon$  & 5 & 45 & 500  &  15.24  \\
32$\epsilon$  & 10 & 110 & 5000  & 18.38 \\
    
\noalign{\hrule height.8pt}
\end{tabular}
\end{table}

\subsection{Adversarial Neuron Manipulation for Multiple Neurons (ANM-M)}\label{subsec:ptbmulti}

ANM-S has shown the potential to exploit pretrained models to generate adversarial perturbations that effectively attack victim models. However, ANM-S focuses on perturbing only a single neuron within the pretrained models. This raises a compelling question: \textit{can stronger attacks be achieved by manipulating multiple neurons simultaneously?}

We begin by conducting a preliminary experiment in which multiple randomly sampled neurons are perturbed. The results reveal that as the number of perturbed neurons increases, the attack’s effectiveness improves significantly compared to perturbing a single neuron (see Section~\ref{sec:exp}). This finding highlights that the collective manipulation of multiple neurons can amplify adversarial perturbations, resulting in a stronger impact on victim models.
Though perturbing random neurons demonstrates some improvement in attack effectiveness, the results lack consistency, likely due to the varying importance of different neurons. Not all neurons contribute equally to the feature representations or the decision-making process of the model, and perturbing less influential neurons may yield negligible adversarial effects. This variation underscores the need for a more targeted approach to neuron selection.

To enhance the effectiveness of \textit{Adversarial Neuron Manipulation (ANM)}, we shift our focus toward identifying neurons that are closely tied to task-specific representations or decision boundaries. By selectively targeting these influential neurons, adversarial effects can be amplified, resulting in more consistent and impactful perturbations.

Recent studies~\cite{jin2020does, jin2022neuronal} analyzing the generalization error of well-trained models have concluded that highly correlated neurons or model weights contribute to generalization error, \ie model degradation. This finding underscores the critical role of neuron importance in the generalizability of deep learning models. Unlike previous works that predominantly investigate neurons or weights at the layer or model level, we narrow the scope to individual neurons and aim to capture their dependencies through the lens of \textit{Mutual Information (MI)}. MI allows us to quantify both \textit{linear} and \textit{non-linear} dependencies, extending beyond the linear correlations considered in~\cite{jin2022neuronal}.
Neurons with higher MI are identified as carriers of significant task-relevant information, making them ideal candidates for adversarial manipulation. However, computing MI for multivariate dependencies is computationally expensive. To address this, we propose an efficient method, termed \textit{Mutual Information Maximization Search (MIMS)}, which employs a greedy and iterative approach to identify neuron candidates. In each iteration, the method seeks to maximize the \textit{incremental mutual information} with the set of neurons already selected. Once the optimal neuron candidates are determined, ANM can be extended to target multiple neurons simultaneously.

The overview of the extended method, termed \textit{ANM-M}, is presented in Fig.~\ref{fig:anm-m}. ANM-M begins by designating a seed neuron \( \phi_0 \), which initializes the selection set \( \Omega = \{\phi_0\} \). We assume neuron activations follow a normal distribution due to the effect of \textit{Batch Normalization (BN)}~\cite{ioffe2015batch}, which explicitly normalizes activations by centering them to zero mean and scaling them to unit variance for each mini-batch. This normalization encourages the activations to approximate a normal distribution, especially in intermediate and deeper layers (see Fig.~\ref{fig:norm_dist}).
Under this assumption, MIMS calculates the mutual information \( I(\Omega; \phi) \) between an existing neuron subset \( \Omega \) and a candidate neuron \( \phi \) with the determinants of their covariance submatrices:
\begin{equation}
I(\Omega; \phi) = \frac{1}{2}\ln\left(\frac{\det(\Sigma_{\Omega})\det(\Sigma_{\phi})}{\det(\Sigma_{\Omega \cup \{\phi\}})}\right),
\end{equation}
where \( \Sigma_{\Omega} \) represents the covariance submatrix of \( \Omega \), \( \Sigma_{\phi} \) is the variance (a \( 1 \times 1 \) submatrix) of the candidate neuron \( \phi \), and \( \Sigma_{\Omega \cup \{\phi\}} \) is the augmented covariance submatrix after incorporating \( \phi \). The candidate neuron that maximizes this MI is added to \( \Omega \), and this process is repeated until the desired number of neurons is selected.

Once the neuron candidates are finalized, the perturbation \( \delta \) for ANM-M is formulated as:
\begin{equation}
\delta = \arg\max_{ \|\delta\|_\infty \leq \epsilon} \sum_{\phi_i \in \Omega} \phi_i(\mathbf{x} + \delta; f_p),
\end{equation}
where \( f_p \) denotes the model's prediction function. To accelerate the optimization process, the final loss function for ANM-M is defined as:
\begin{equation}
\mathcal{L}_{\text{ANM-M}} = \mathbb{E}_{\mathbf{x} \in \mathcal{D}, \phi_i \in \Omega} \left[ \left(t_i - \phi_i\left(\mathbf{x} + \delta; f_p\right)\right)^2\right],
\end{equation}
where \( t \) is a hyperparameter controlling the intensity of the adversarial attack.
This approach not only enhances the scalability of ANM but also improves the effectiveness of adversarial perturbations by leveraging a targeted selection of influential neurons. 
The algorithm of ANM-M can be found in Algorithm.~\ref{algo:anm-m}.

\section{Experiments}\label{sec:exp}
In this section, we evaluate the effectiveness of the proposed method through comprehensive experiments. The evaluation covers diverse datasets, carefully designed implementation details, and a thorough analysis of the results.

\begin{figure*}[t]
    \begin{minipage}[t]{0.3\textwidth}
        \centering
        \vspace{0pt}
        \includegraphics[width=0.8\linewidth]{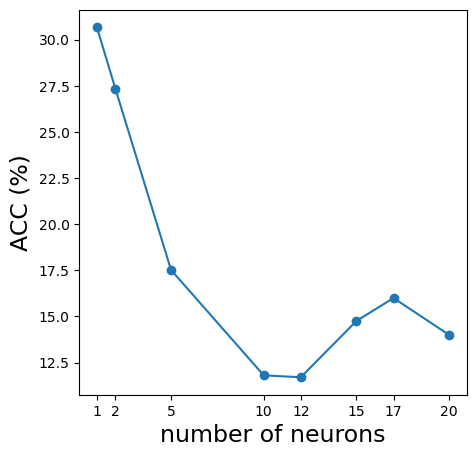}
        \captionof{figure}{Classification Accuracy of ResNet18 under ANM-M attack with different number of neurons.}
        \label{fig:num_neuron}
    \end{minipage}%
    \hfill
    \begin{minipage}[t]{0.65\textwidth}
        \vspace{-6pt}
        \captionof{table}{Classification Accuracy (\%) under Our Attacks. The absolute accuracy drop from clean accuracy is indicated in \textcolor{red}{red}(the higher, the better). The best attack is highlighted in \textbf{BOLD}(The lower, the better).}
        \label{table:results}
        \resizebox{\linewidth}{!}{%
        \begin{tabular}{cccccc}
\noalign{\hrule height.8pt}
Dataset       & \begin{math} f_d \end{math} & ACC(\%) & ANM-S (\%)  & ANM-RANDOM (\%) & ANM-M (\%)\\
\noalign{\hrule height.7pt}
\multirow{4}*{\makecell{UCM}}  & ResNet18     & 93.24   & 18.28   \textcolor{red}{($\downarrow$74.96)}  & 11.81       \textcolor{red}{($\downarrow$81.43)} & \textbf{11.49} \textcolor{red}{($\downarrow$81.75)}
\\
& ResNet50     & 94.48   & 20.91 \textcolor{red}{($\downarrow$73.57)}   & 16.38  \textcolor{red}{($\downarrow$78.10)} & \textbf{14.35}
\textcolor{red}{($\downarrow$80.13)}\\
    & VGG11     & 90.76   & 50.18    \textcolor{red}{($\downarrow$40.58)}  & 32.29      \textcolor{red}{($\downarrow$58.47)}  & \textbf{28.28} \textcolor{red}{($\downarrow$62.48)}
\\
    & VGG19     & 91.33   & 28.39     \textcolor{red}{($\downarrow$62.94)}  & 18.10     \textcolor{red}{($\downarrow$73.23)}  &\textbf{16.84} \textcolor{red}{($\downarrow$79.49)}
 \\
    \hhline{~-----}
    \multicolumn{2}{r}{Average Drop} & - &\textcolor{red}{$\downarrow$63.01} & \textcolor{red}{$\downarrow$72.81}&\textcolor{red}{ $\downarrow$75.96}
    \\

\noalign{\hrule height.6pt}
\multirow{4}*{\makecell{AID}} 
 & ResNet18     & 89.66   & \ 7.14     \textcolor{red}{($\downarrow$82.52)}  & \ 8.68   \textcolor{red}{($\downarrow$80.98)}  & \textbf{\ 5.42} \textcolor{red}{($\downarrow$84.24)}
\\
& ResNet50     & 90.44   & 10.93  \textcolor{red}{($\downarrow$79.51)}     & \ 8.18  \textcolor{red}{($\downarrow$82.26)}   &  \textbf{\ 6.82} \textcolor{red}{($\downarrow$83.62)}
\\

    & VGG11     & 89.80   & 35.85    \textcolor{red}{($\downarrow$53.95)}  &19.24    \textcolor{red}{($\downarrow$70.56)}   & \textbf{16.14} \textcolor{red}{($\downarrow$73.66)}
 \\
    & VGG19     & 87.40   & 24.42  \textcolor{red}{($\downarrow$62.98)}  & 11.21      \textcolor{red}{($\downarrow$76.19)}  &  \textbf{\ 8.93} \textcolor{red}{($\downarrow$78.47)}
 \\
    \hhline{~-----}
     &Average Drop & - &\textcolor{red}{$\downarrow$69.74} & \textcolor{red}{$\downarrow$77.50}&\textcolor{red}{ $\downarrow$80.00}
    \\
\noalign{\hrule height.8pt}
\end{tabular}
}
    \end{minipage}
\end{figure*}

\begin{figure}[t]
    \centering
    \begin{subfigure}[t]{0.24\columnwidth}
        \centering
        \includegraphics[width=\linewidth]{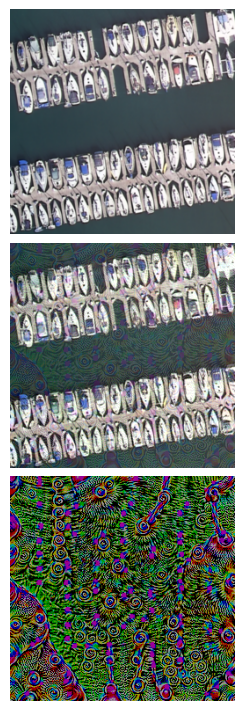}
        \caption{ResNet50}
        \label{fig:ptb-1-ucm_a}
    \end{subfigure}
    \hspace*{-10pt}
    \begin{subfigure}[t]{0.24\columnwidth}
        \centering
        \includegraphics[width=\linewidth]{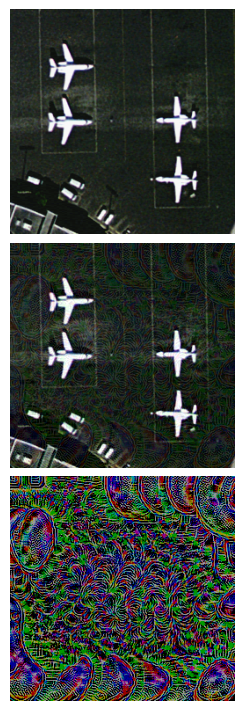}
        \caption{ResNet18}
        \label{fig:ptb-1-ucm_b}
    \end{subfigure}
    \hspace*{-10pt}
    \begin{subfigure}[t]{0.24\columnwidth}
        \centering
        \includegraphics[width=\linewidth]{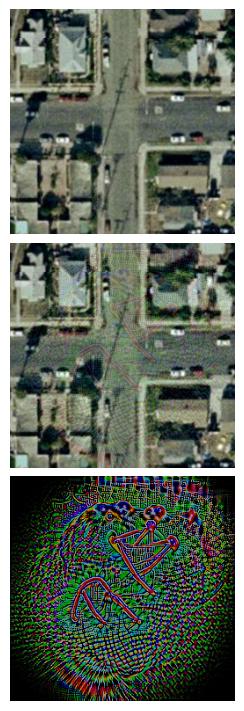}
        \caption{VGG19}
        \label{fig:ptb-1-ucm_c}
    \end{subfigure}
    \hspace*{-10pt}
    \begin{subfigure}[t]{0.24\columnwidth}
        \centering
        \includegraphics[width=\linewidth]{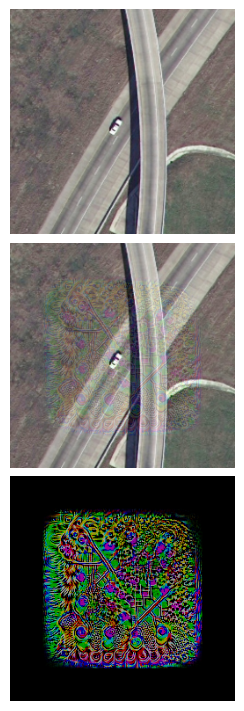}
        \caption{VGG11}
        \label{fig:ptb-1-ucm_d}
    \end{subfigure}
    
    \caption{
        Examples of adversarial perturbations generated by ANM-S and applied to the UCM dataset. Each column corresponds to a different feature extractor model (ResNet50, ResNet18, VGG19, and VGG11). From top to bottom, there are original clean images, adversarial examples and adversarial perturbations generated on ImageNet (amplified 10x for visibility). 
    }
    \label{fig:ptb-1-ucm}
\end{figure}

\begin{figure}[t]
    \centering
    \begin{subfigure}[t]{0.45\columnwidth}
        \centering
        \includegraphics[width=\linewidth]{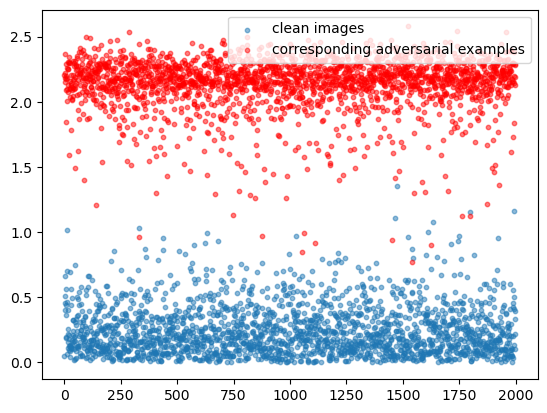}
        \caption{Neuron $2015^{th}$ (ImageNet)}
        \label{fig:neuron_value_vis_a}
    \end{subfigure}
    \begin{subfigure}[t]{0.45\columnwidth}
        \centering
        \includegraphics[width=\linewidth]{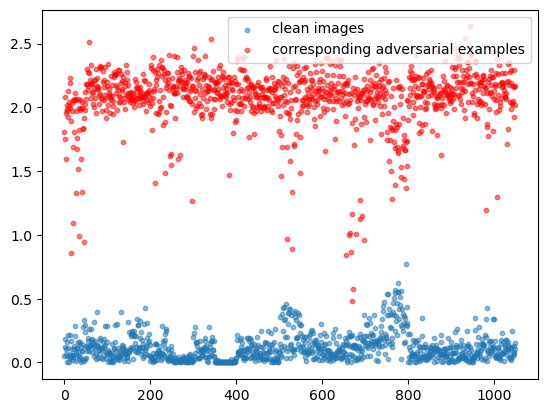}
        \caption{Neuron $2015^{th}$ (UCM)}
        \label{fig:neuron_value_vis_b}
    \end{subfigure}
    
    \vspace{5pt} 

    \begin{subfigure}[t]{0.45\columnwidth}
        \centering
        \includegraphics[width=\linewidth]{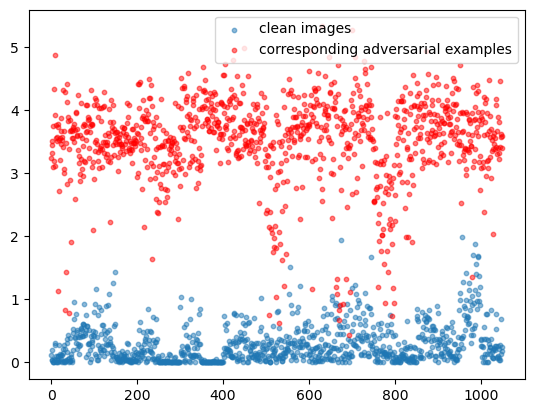}
        \caption{Neuron $1640^{th}$ (UCM)}
        \label{fig:neuron_value_vis_c}
    \end{subfigure}
    \begin{subfigure}[t]{0.45\columnwidth}
        \centering
        \includegraphics[width=\linewidth]{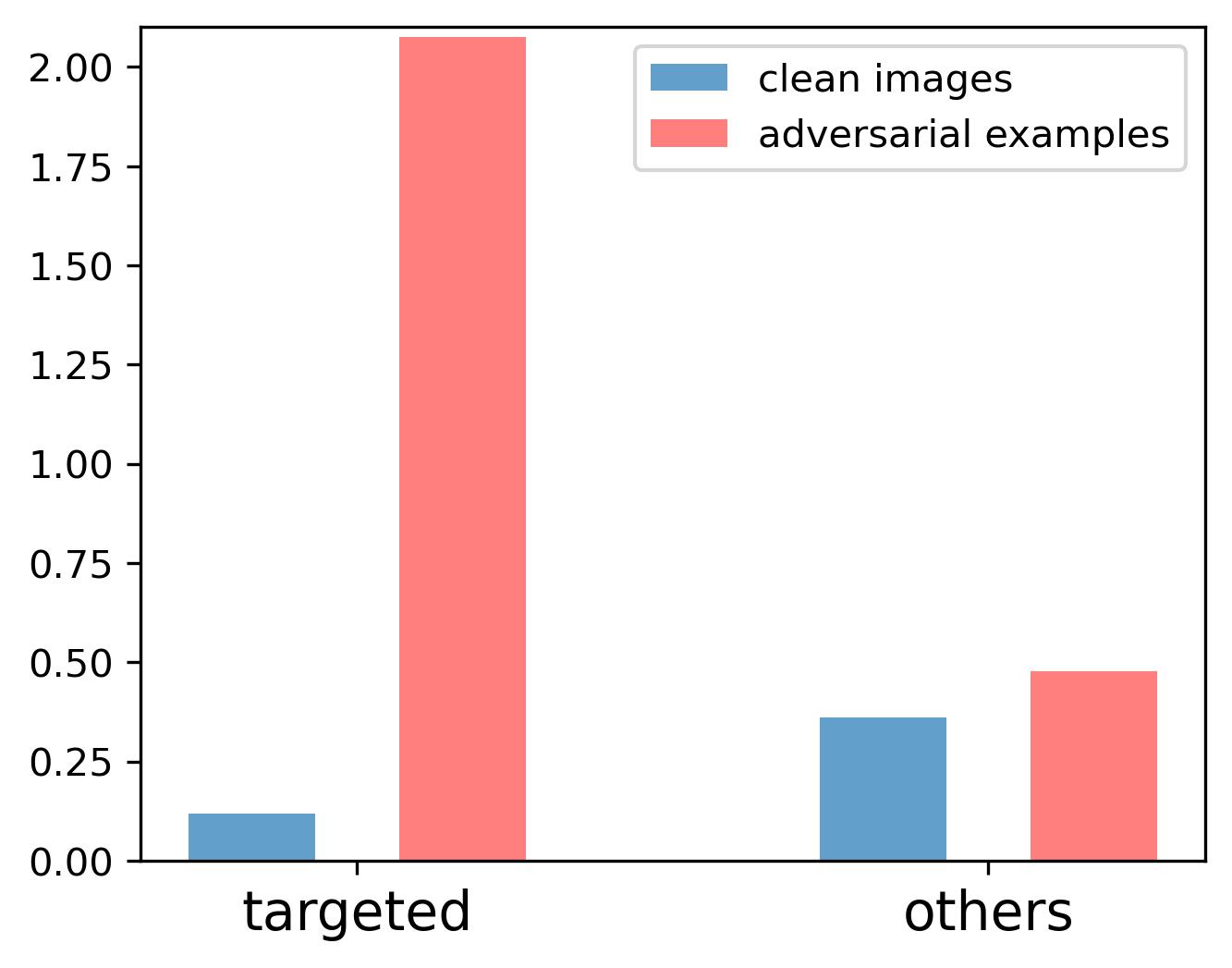}
        \caption{Targeted vs Others (UCM)}
        \label{fig:comp}
    \end{subfigure}
    
    \caption{
        Neuron values on \textcolor{blue}{clean samples} and \textcolor{red}{adversarial examples}. We use ResNet50 here as the pretrained and downstream model.
    }
    \label{fig:neuron_value_vis}
\end{figure}

\subsection{Experiment Settings}\label{subsec:setting}

\paragraph{Datasets}
For the training of the downstream models and the evaluation of 
ANM attack performance, we use two popular remote sensing datasets in our experiments: UCM \cite{yang2010bag} and AID \cite{xia2017aid}.
UCM dataset consists of 21 classes, each containing 100 aerial images with a resolution of $256 \times 256$ pixels. The classes include agricultural, airplane, baseball diamond, beach, buildings, chaparral, dense residential, forest, freeway, golf course, harbor, intersection, medium-density residential, mobile home park, overpass, parking lot, river, runway, sparse residential, storage tanks, and tennis courts. The spatial resolution of the images is 0.3 meters per pixel. The dataset is split equally, with 5,000 images allocated for training and 5,000 for testing.
The AID dataset comprises 30 classes with a total of 10,000 images, evenly split into 5,000 images for training and 5,000 for testing, each sized $600 \times 600$ pixels. The number of samples per class ranges from 200 to 400, with spatial resolutions varying from 0.5 m to 8 m per pixel. The classes include airport, bare land, baseball field, beach, bridge, center, church, commercial, dense residential, desert, farmland, forest, industrial, meadow, medium-density residential, mountain, park, parking lot, playground, pond, port, railway station, resort, river, school, sparse residential, square, stadium, storage tanks, and viaduct. 

The generation dataset refers to the publicly available dataset $\mathcal{D}$ used during the perturbation generation process described in Algorithm \ref{algo:anm-s} and Algorithm \ref{algo:anm-m}. As outlined in Section~\ref{subsec:preli}, no information regarding downstream classification tasks is accessible; therefore, the generation dataset cannot be task-specific, \eg remote sensing dataset. Instead, 5,000 images were randomly sampled from the ImageNet-1K validation set\footnote{https://image-net.org/challenges/LSVRC/2012/2012-downloads.php} to serve as the generation dataset.

\paragraph{Victim Models}
We mainly use two representative types of deep neural networks: VGG~\cite{simonyan2014very} and ResNet~\cite{he2016deep} in this paper.
Specifically, we have four models: VGG11, VGG18, ResNet18 and ResNet50 pretrained on ImageNet.
Note that original VGG~\cite{simonyan2014very} doesn't have Batch Normalization, but Batch Normalization has been integrated with VGG in implementation by popular deep learning frameworks, \eg Pytorch\footnote{https://pytorch.org/vision/main/models/vgg.html}.
As described in Section \ref{subsec:preli}, our goal is to reveal the adversarial risks of transfer learning and the victim models are actually finetuned counterparts of above models.
For finetuning, we freeze most parameters of the pretrained models and only retrain the last (few) layers. Specifically, for ResNets, the last one fully connected layers are retrained, while for VGG models, retraining the last three fully connected layers yields satisfactory classification accuracy. The transfer learning configuration employs a learning rate of $1\times10^{-4}$
 , with training conducted over 50 epochs using minibatches of size 10. 




\begin{figure*}[t]
    \centering
    \begin{subfigure}[t]{0.24\textwidth}
        \centering
        \includegraphics[width=\linewidth]{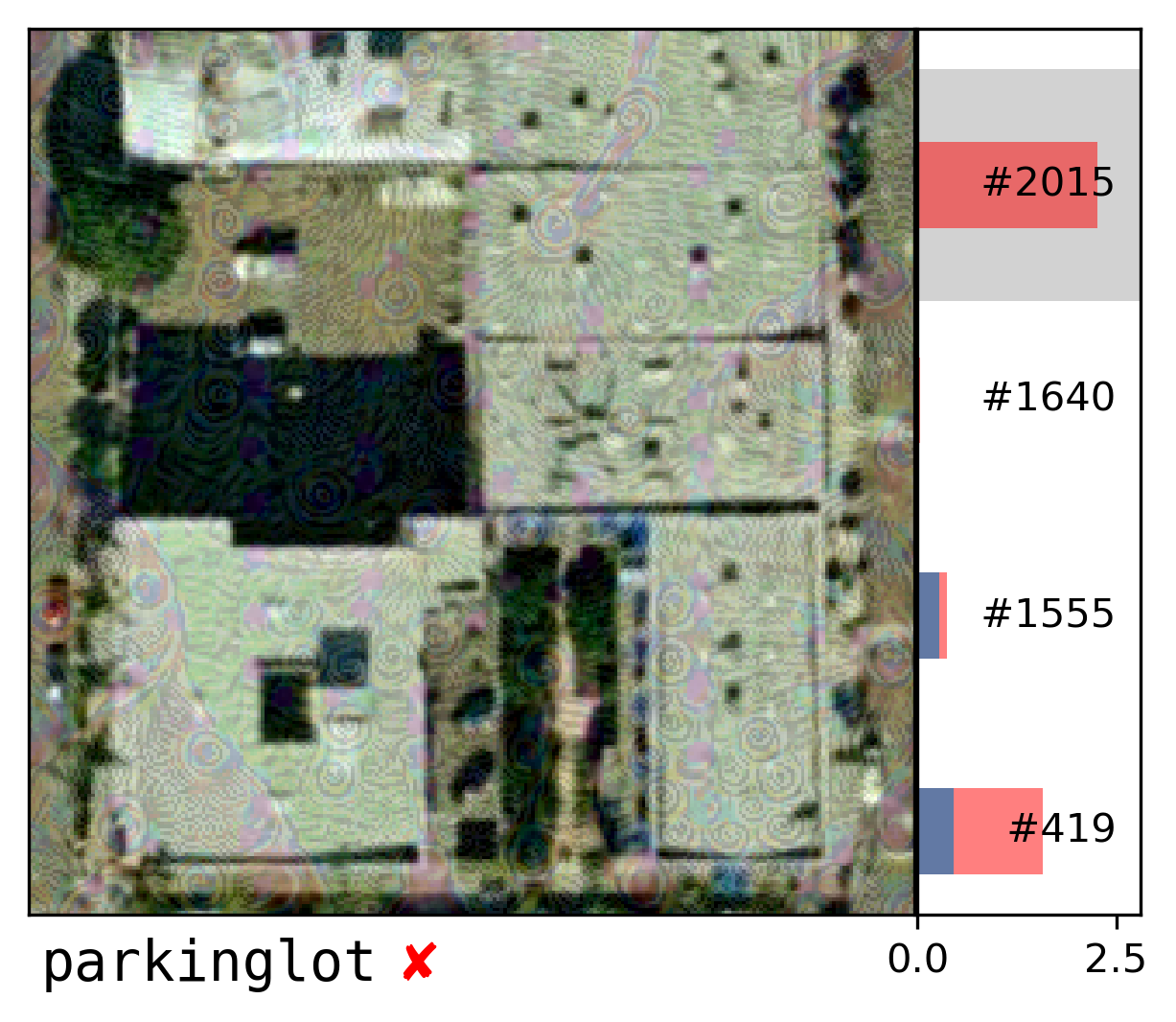}
        \label{fig:ptb1a}
    \end{subfigure}
    \hfill
    \begin{subfigure}[t]{0.24\textwidth}
        \centering
        \includegraphics[width=\linewidth]{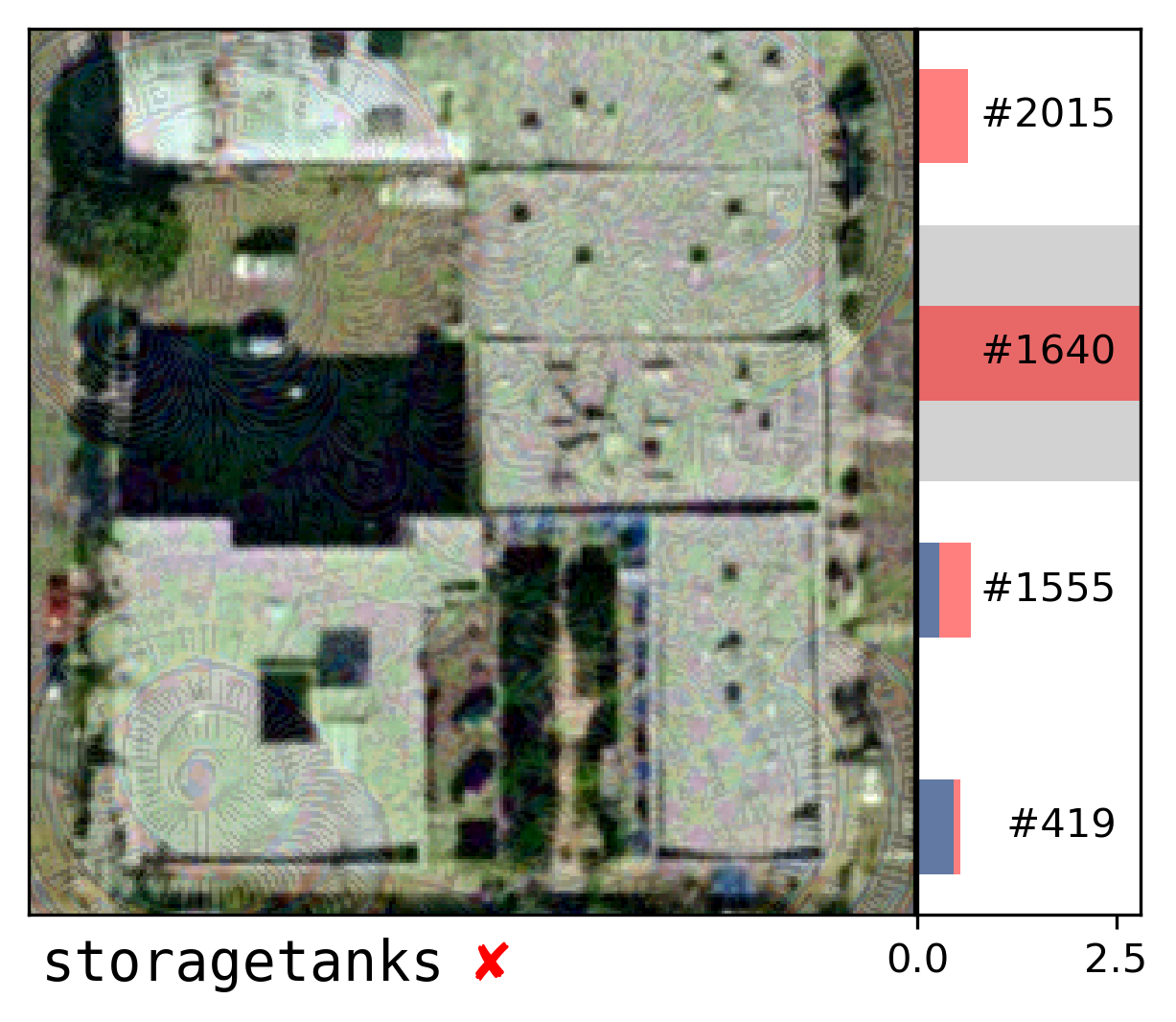}
        \label{fig:ptb1b}
    \end{subfigure}
    \hfill
    \begin{subfigure}[t]{0.24\textwidth}
        \centering
        \includegraphics[width=\linewidth]{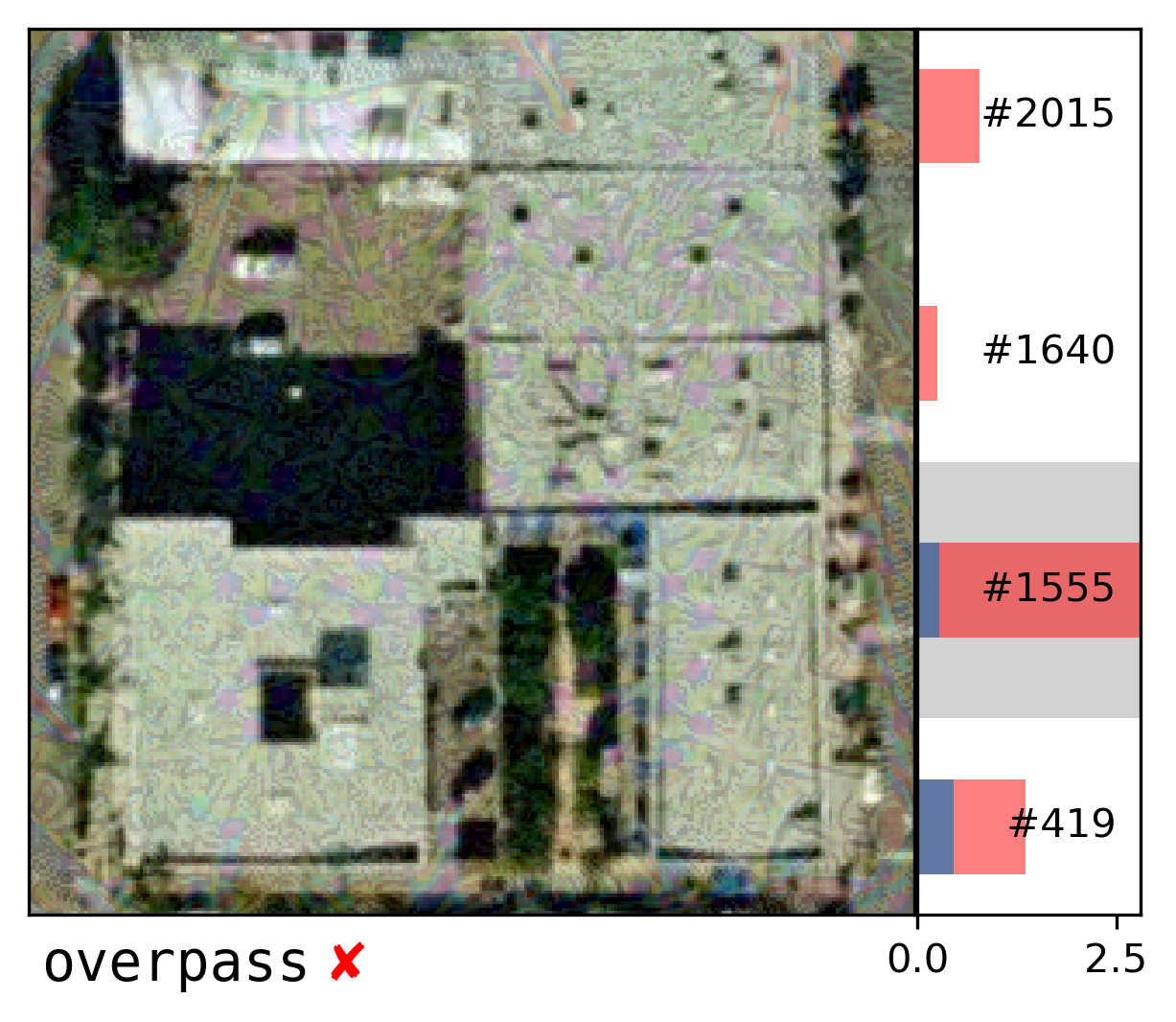}
        \label{fig:ptb1c}
    \end{subfigure}
    \hfill
    \begin{subfigure}[t]{0.24\textwidth}
        \centering
        \includegraphics[width=\linewidth]{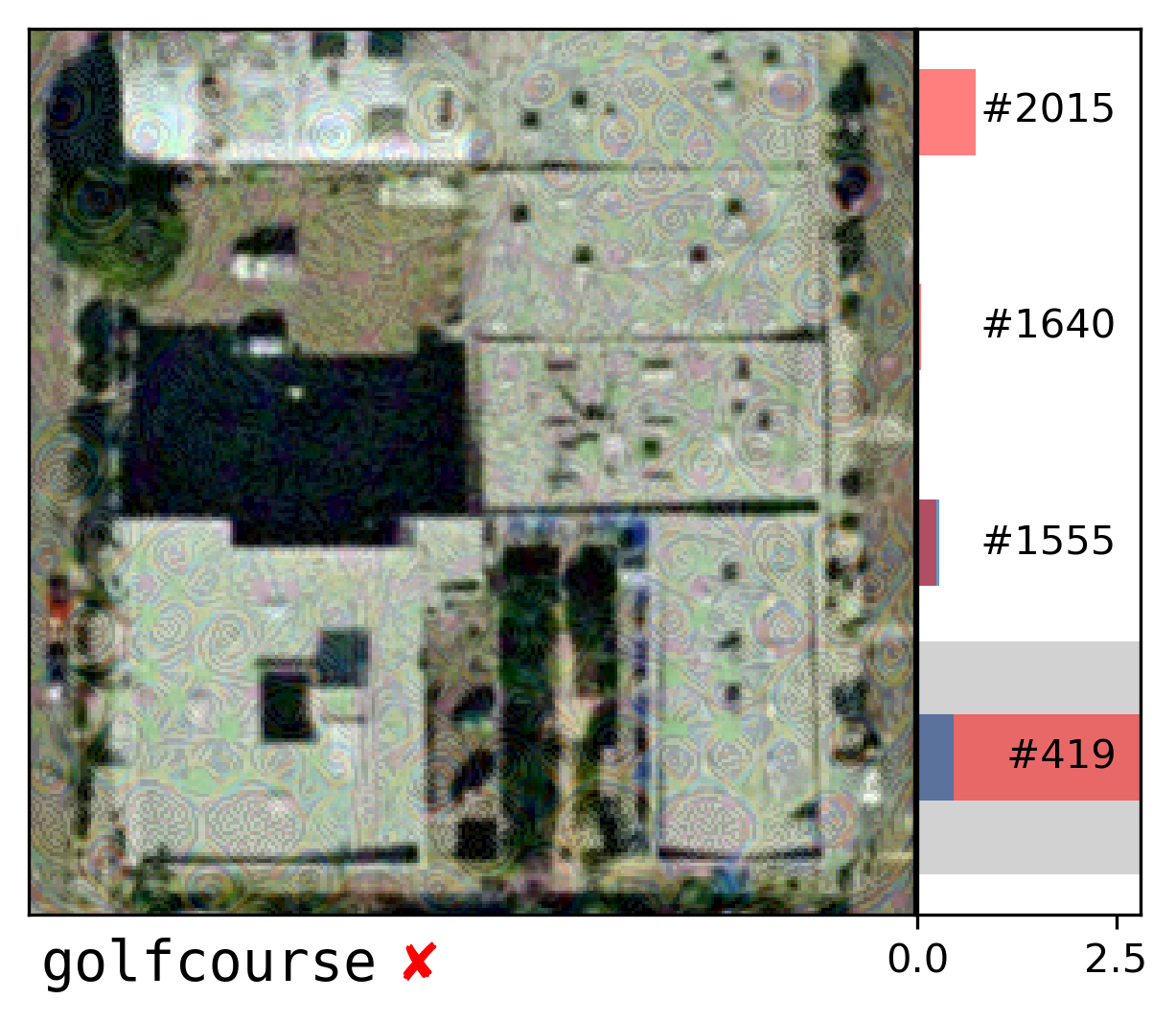}
        \label{fig:ptb1d}
    \end{subfigure}
    
    \vspace{-1em}
    \begin{subfigure}[t]{0.24\textwidth}
        \centering
        \includegraphics[width=\linewidth]{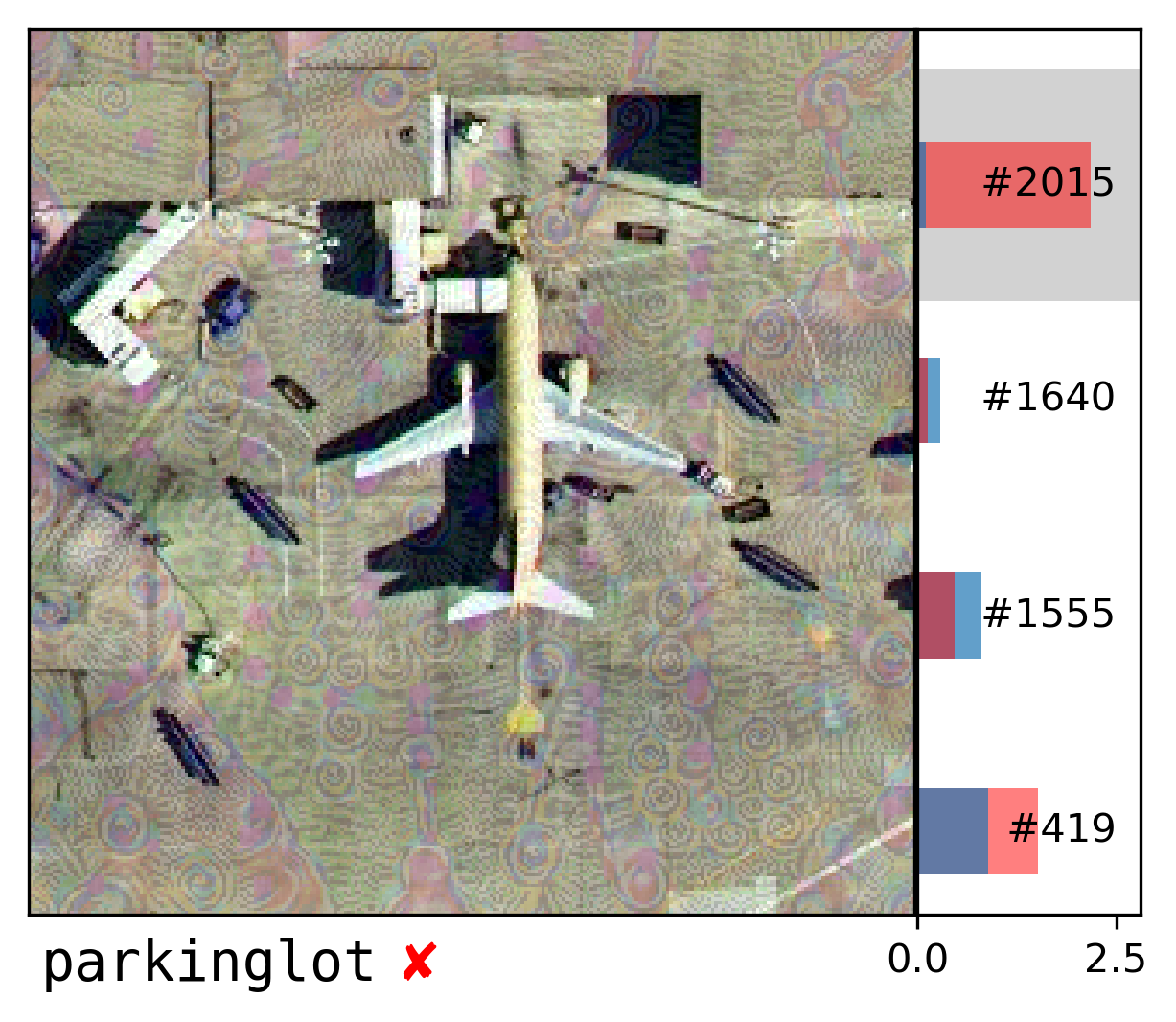}
        \label{fig:ptb2a}
    \end{subfigure}
    \hfill
    \begin{subfigure}[t]{0.24\textwidth}
        \centering
        \includegraphics[width=\linewidth]{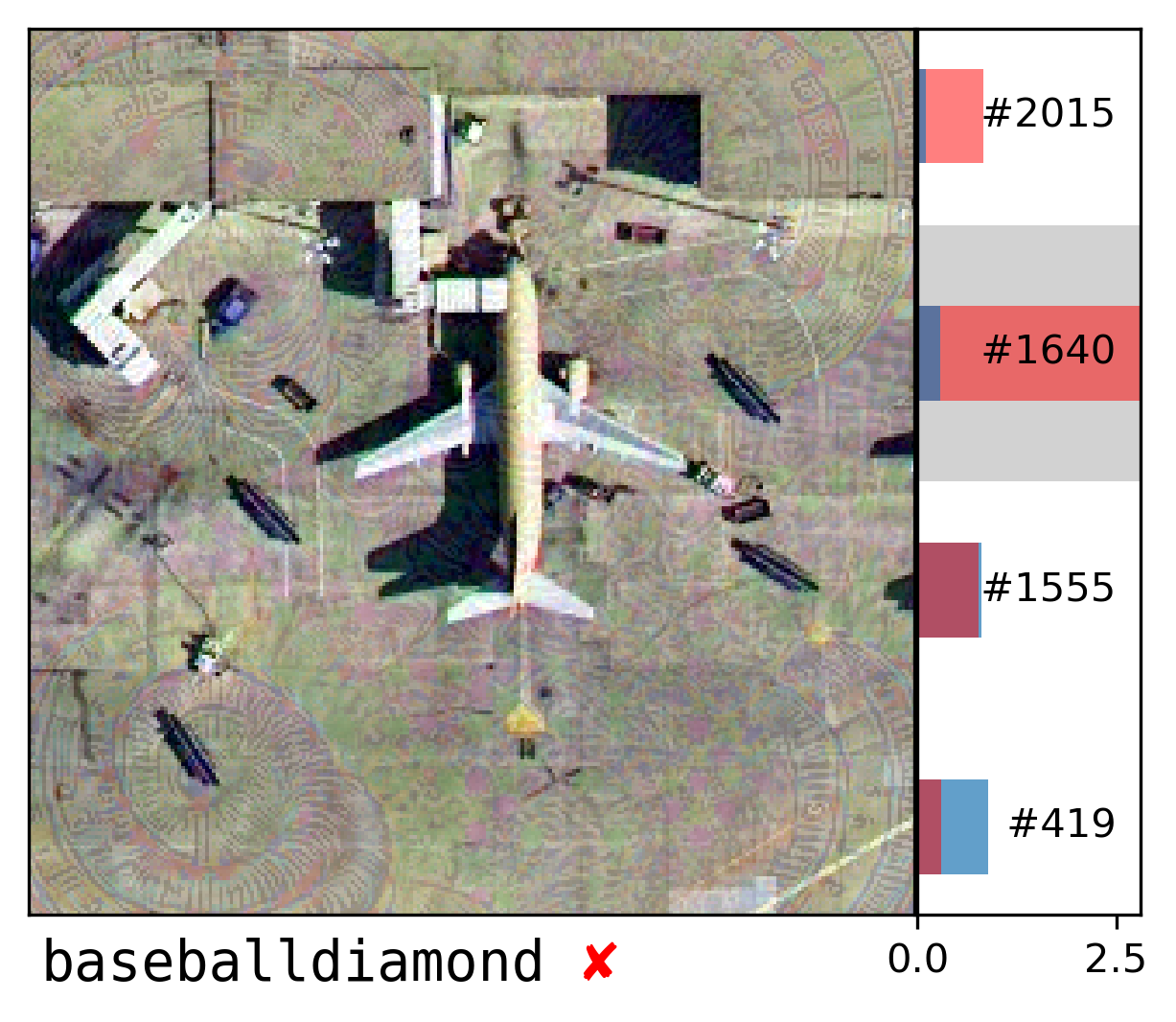}
        \label{fig:ptb2b}
    \end{subfigure}
    \hfill
    \begin{subfigure}[t]{0.24\textwidth}
        \centering
        \includegraphics[width=\linewidth]{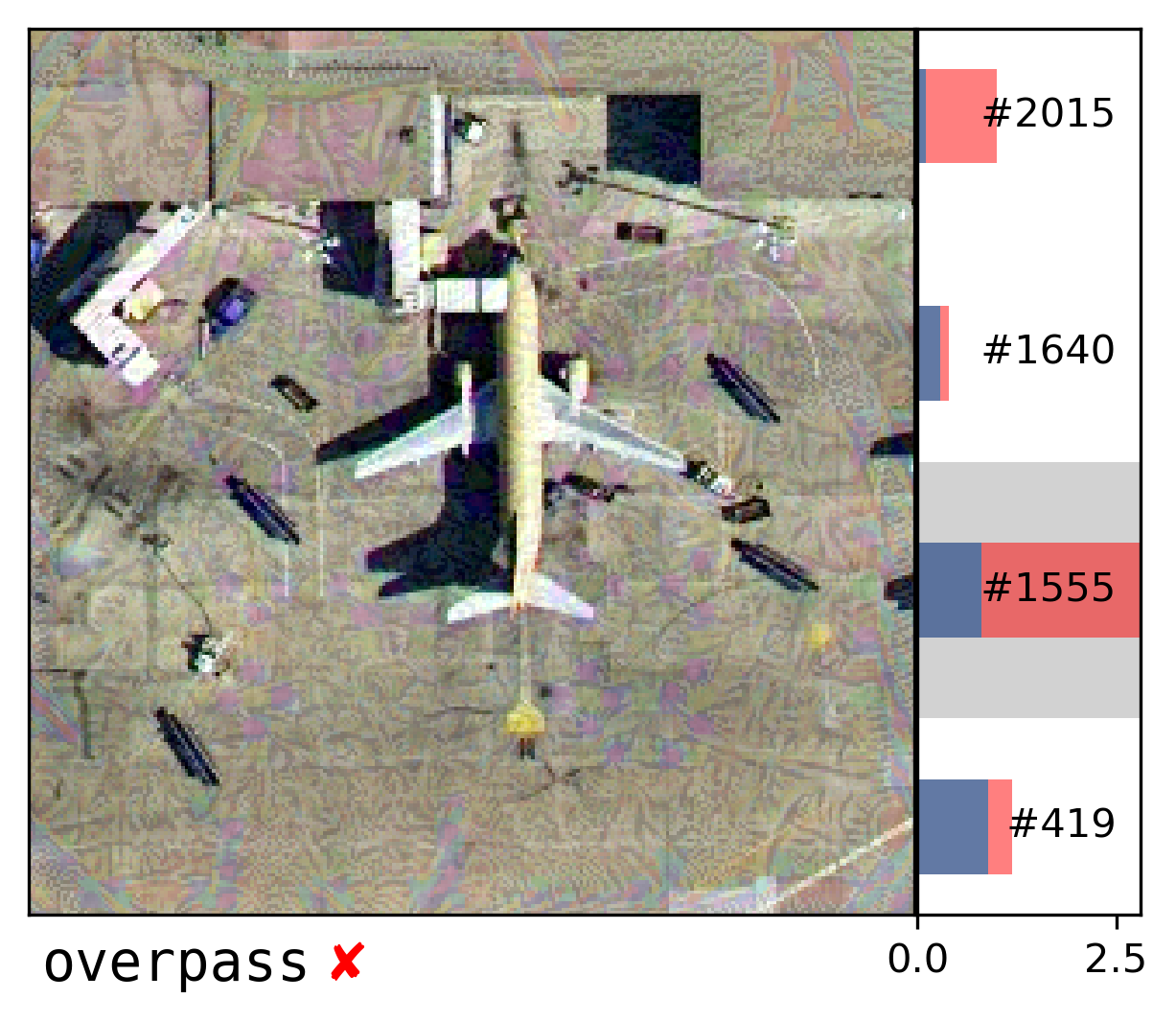}
        \label{fig:ptb2c}
    \end{subfigure}
    \hfill
    \begin{subfigure}[t]{0.24\textwidth}
        \centering
        \includegraphics[width=\linewidth]{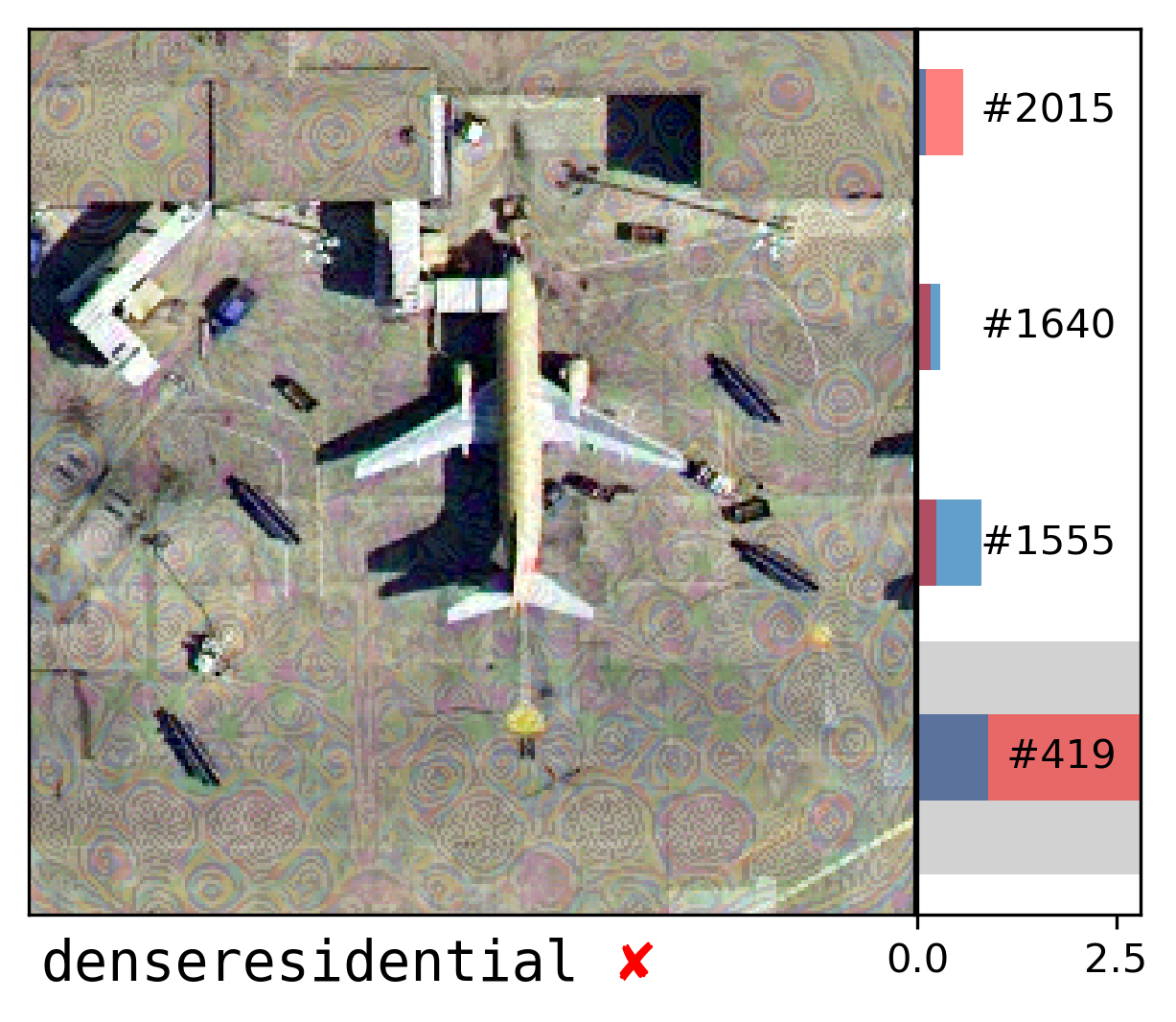}
        \label{fig:ptb2d}
    \end{subfigure}
    
    
    \caption{
       Amplification of neuron values caused by ANM-S. Target neuron is indicated with grey shadow.
       From left to right, the target neurons are 2015th, 1640th, 1555th and 419th neuron of ResNet50. Blue bars are the neuron values of clean input images and red are those of the adversarial examples. The labels below represent the predictions made by a transfer-learned ResNet50 model when applied to adversarial examples. The correct predictions for the first and second rows should be \textit{buildings} and \textit{airplane}, respectively.
    }
    \label{fig:anm-s-values}
\end{figure*}

\begin{figure*}[h]
    \centering
    \begin{subfigure}[t]{0.24\textwidth}
        \centering
        \includegraphics[width=\linewidth]{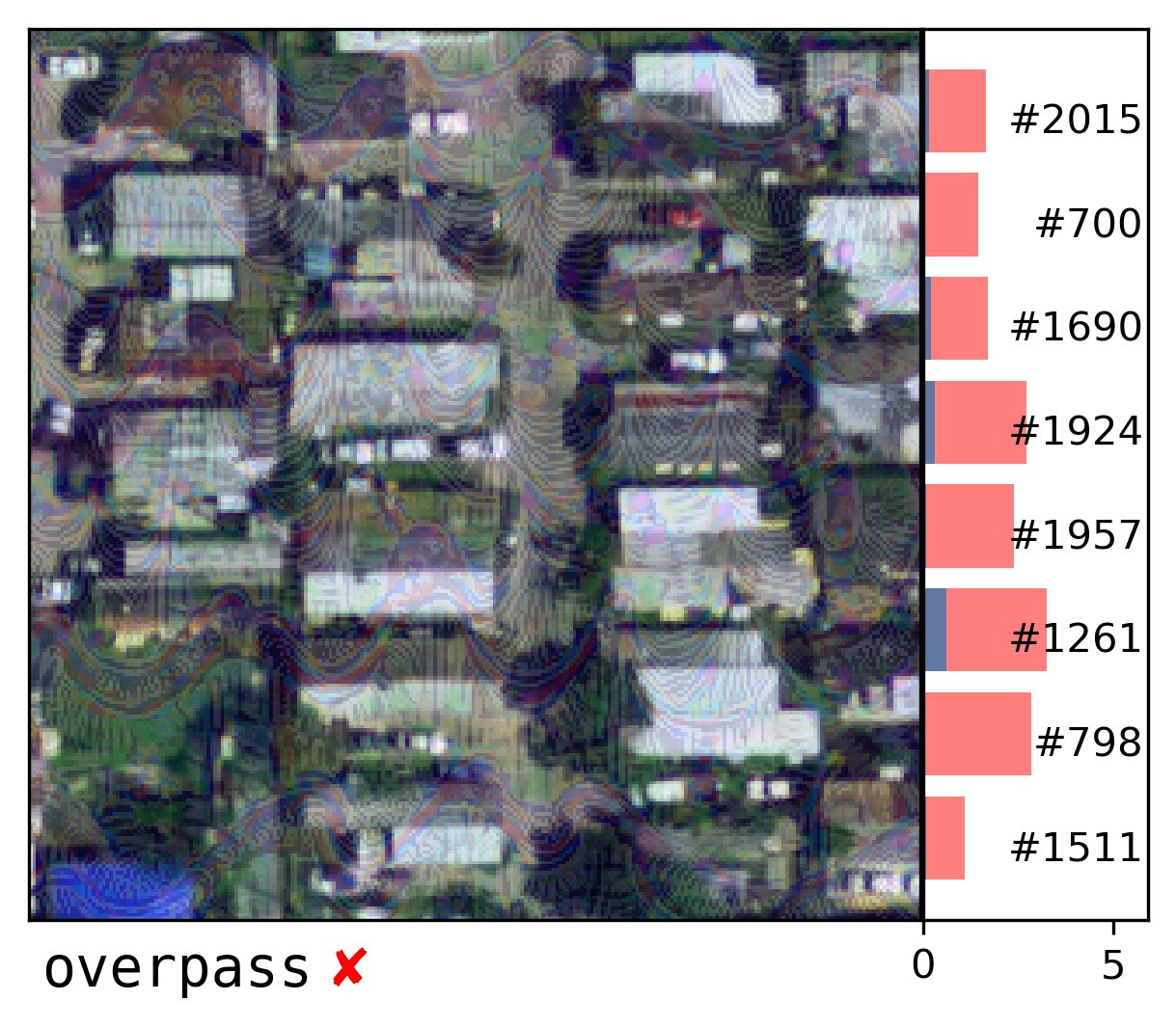}
        \label{fig:ptb-multi_a}
    \end{subfigure}
    \hfill
    \begin{subfigure}[t]{0.24\textwidth}
        \centering
        \includegraphics[width=\linewidth]{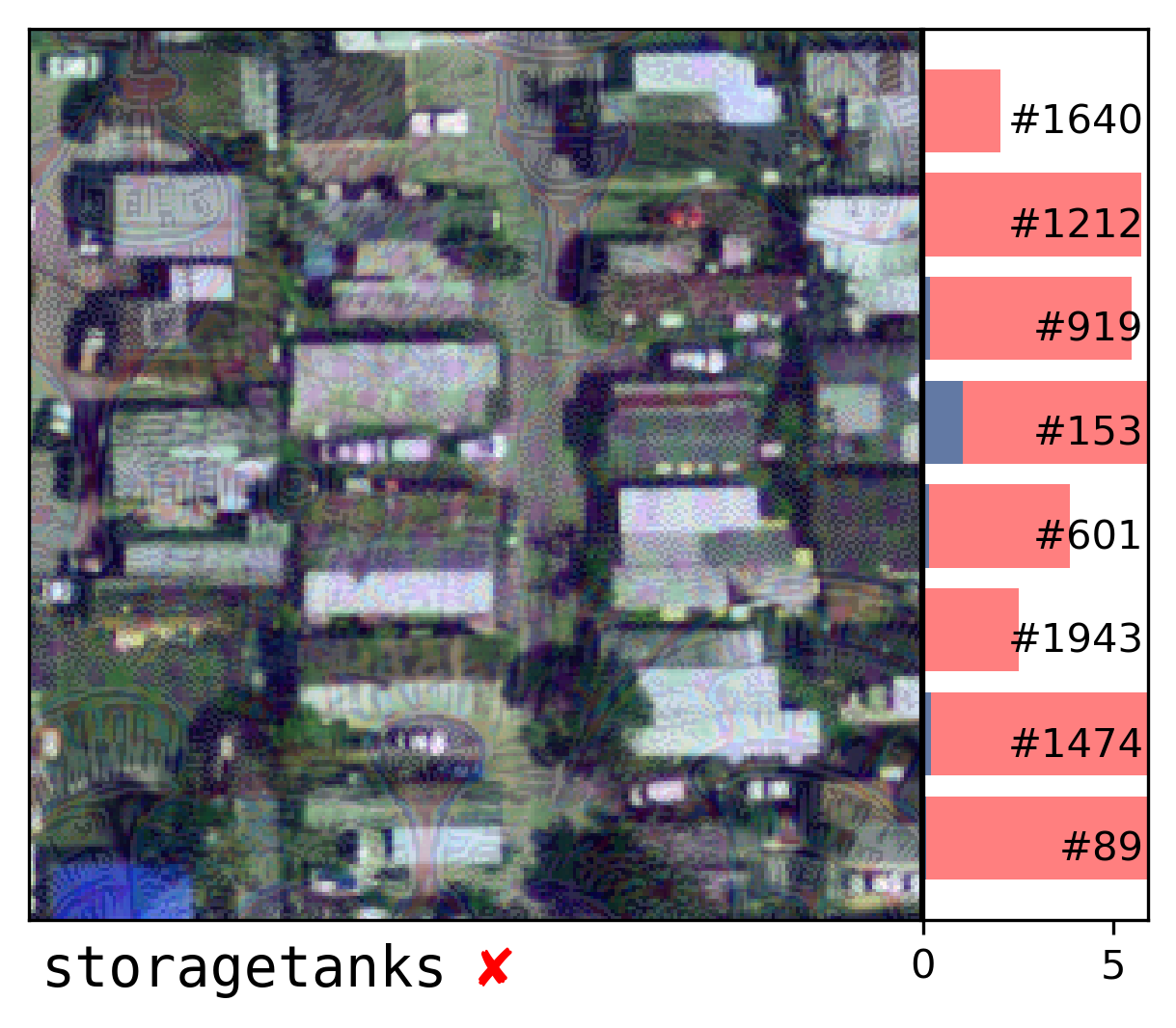}
        \label{fig:ptb-multi_b}
    \end{subfigure}
    \hfill
    \begin{subfigure}[t]{0.24\textwidth}
        \centering
        \includegraphics[width=\linewidth]{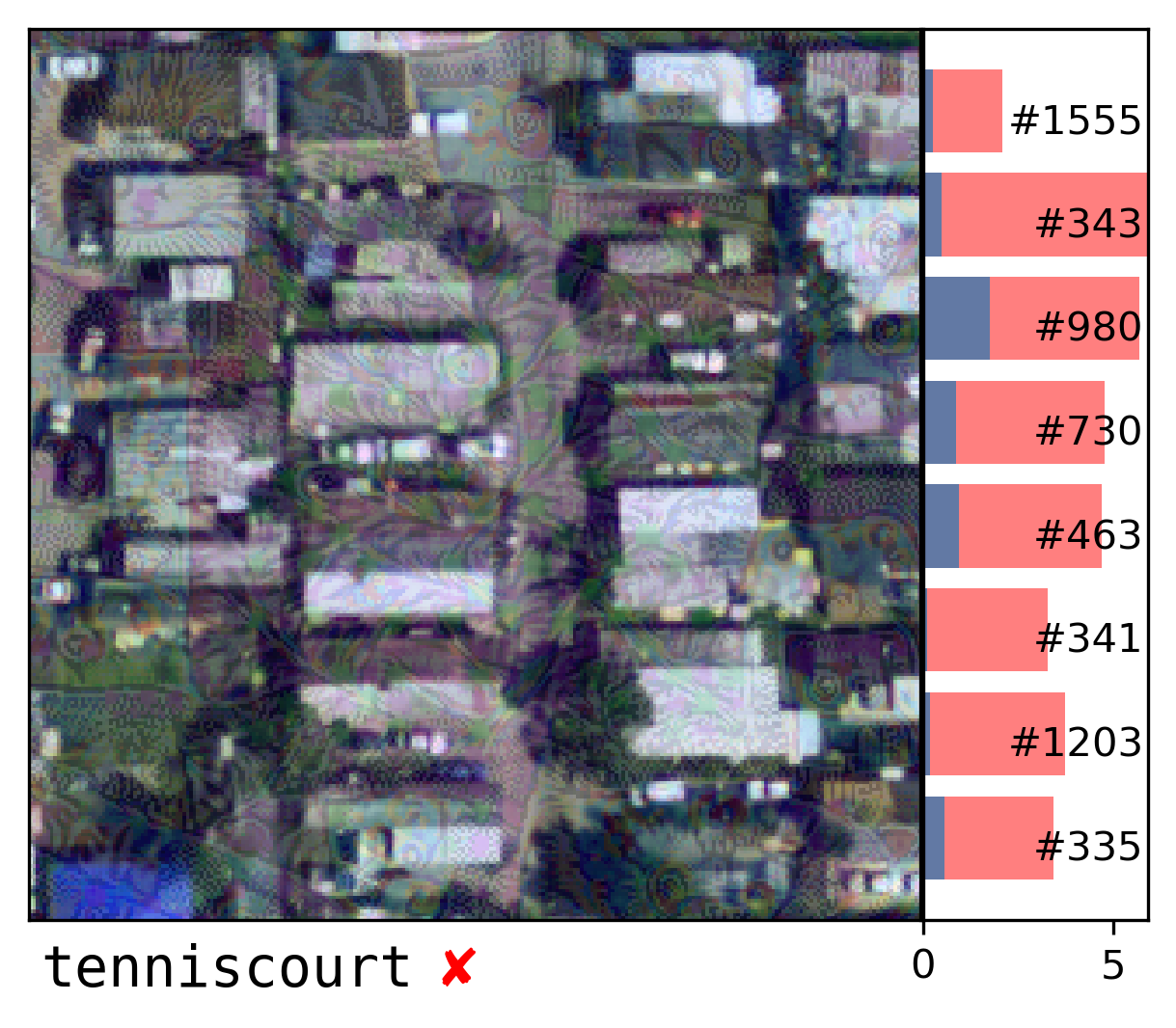}
        \label{fig:ptb-multi_c}
    \end{subfigure}
    \hfill
    \begin{subfigure}[t]{0.24\textwidth}
        \centering
        \includegraphics[width=\linewidth]{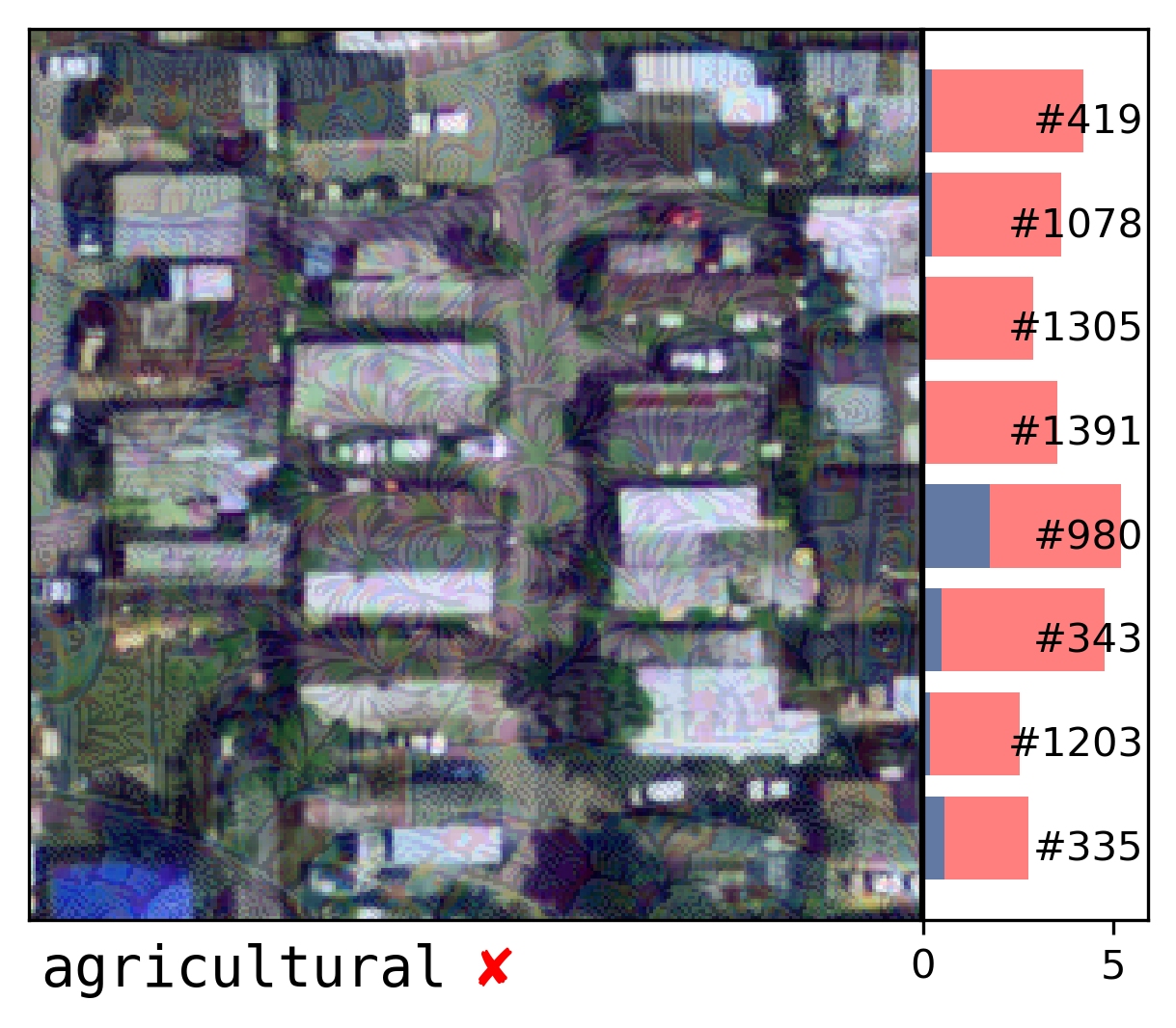}
        \label{fig:ptb-multi_d}
    \end{subfigure}
    
    \vspace{-1em} 
    \begin{subfigure}[t]{0.24\textwidth}
        \centering
        \includegraphics[width=\linewidth]{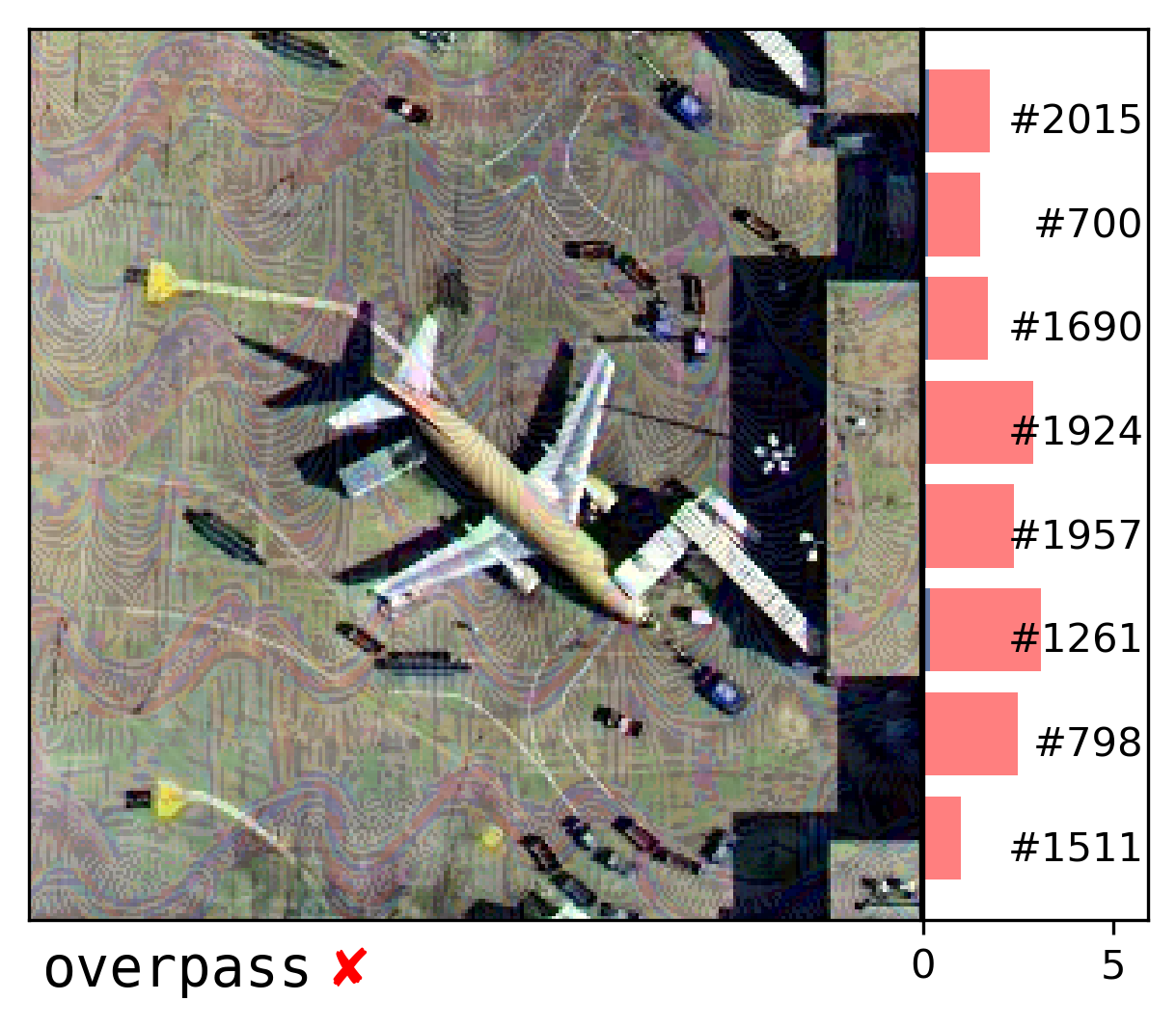}
        \label{fig:ptb-multi_e}
    \end{subfigure}
    \hfill
    \begin{subfigure}[t]{0.24\textwidth}
        \centering
        \includegraphics[width=\linewidth]{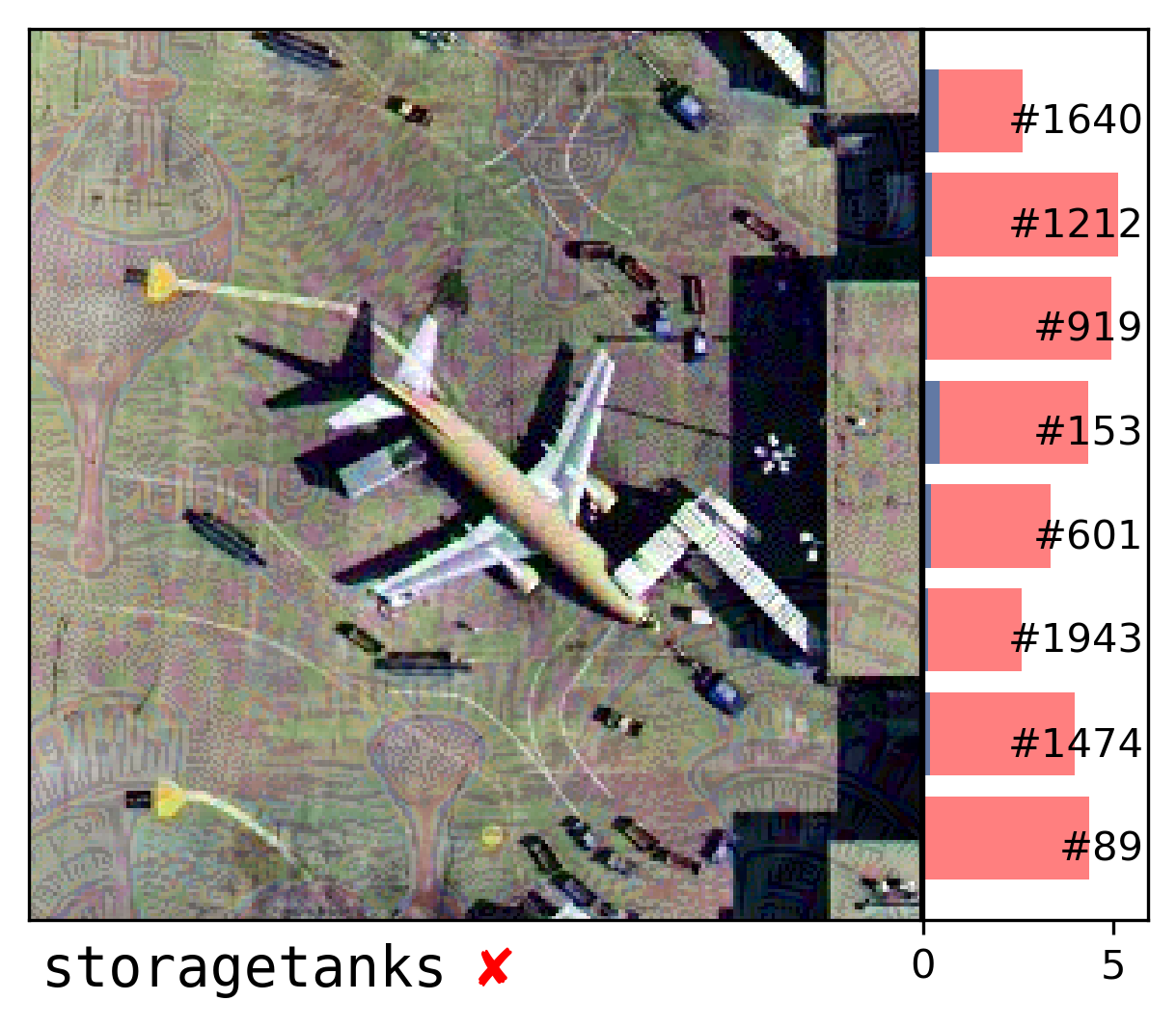}
        \label{fig:ptb-multi_f}
    \end{subfigure}
    \hfill
    \begin{subfigure}[t]{0.24\textwidth}
        \centering
        \includegraphics[width=\linewidth]{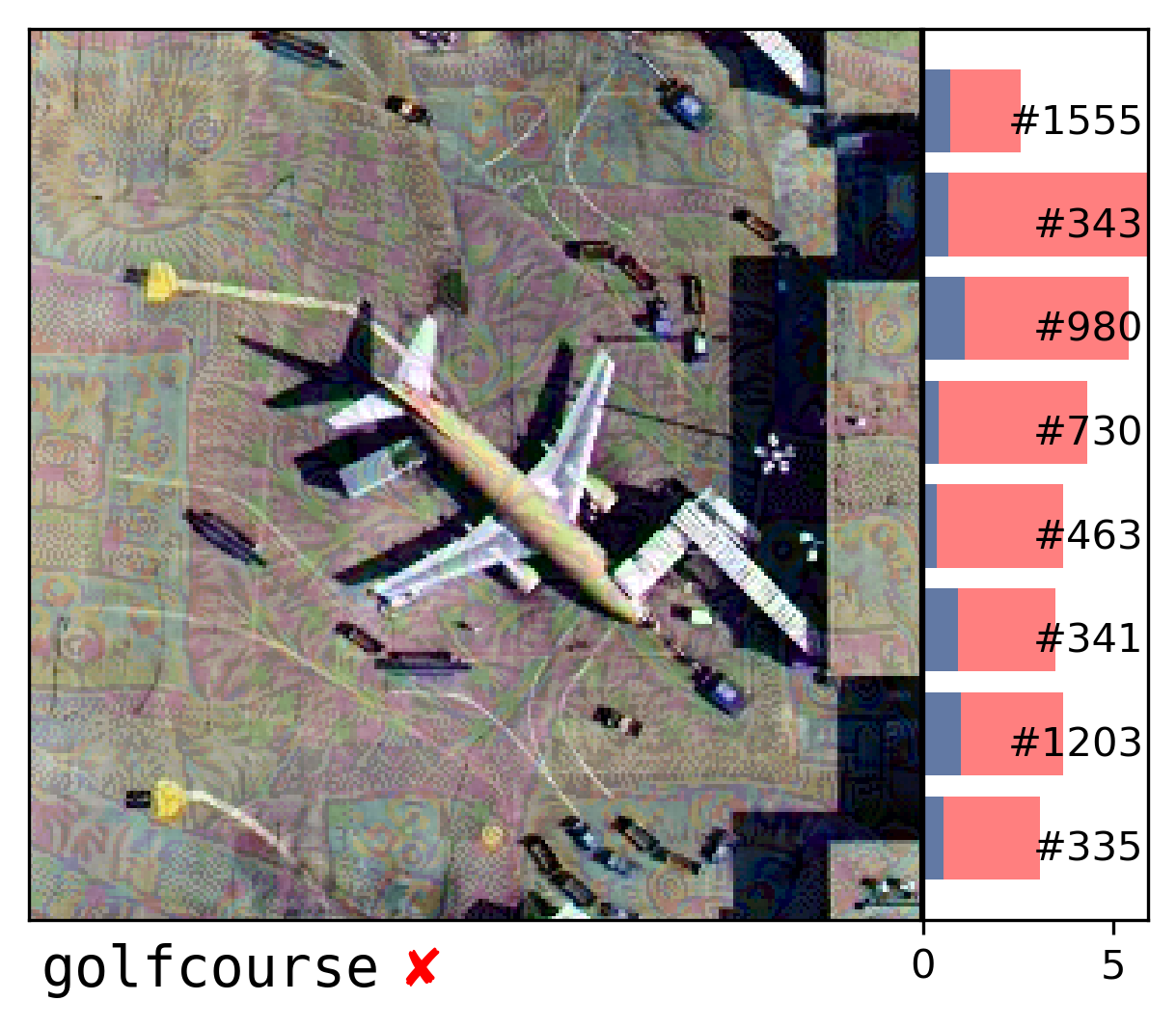}
        \label{fig:ptb-multi_g}
    \end{subfigure}
    \hfill
    \begin{subfigure}[t]{0.24\textwidth}
        \centering
        \includegraphics[width=\linewidth]{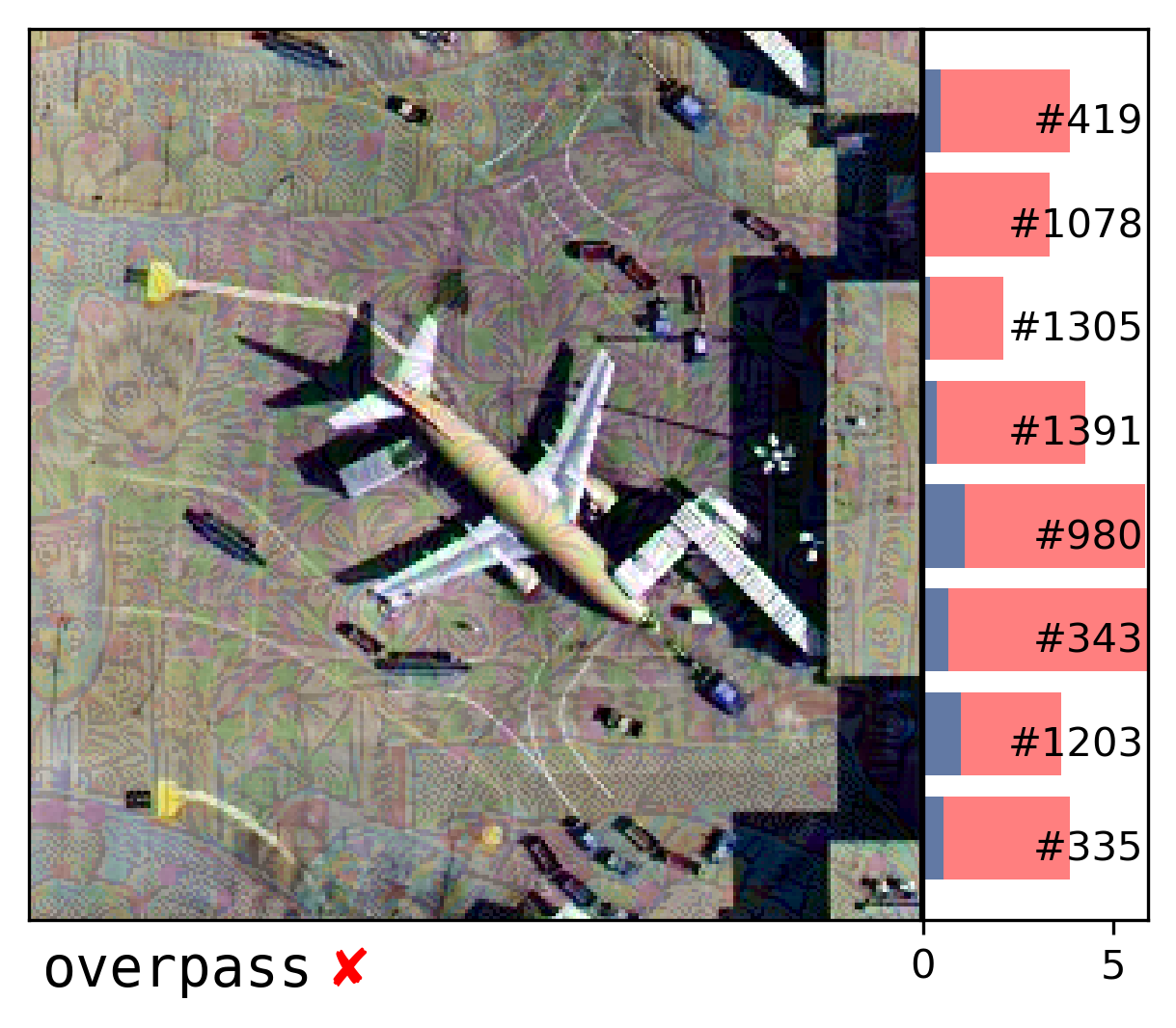}
        \label{fig:ptb-multi_h}
    \end{subfigure}
    
    
    \caption{
        Amplification of neuron values caused by ANM-M. Blue bars are the neuron values of clean input images and red are those of the adversarial examples. The labels below represent the predictions made by a transfer-learned ResNet50 model when applied to adversarial examples. The correct predictions for the first and second rows are \textit{mobilehomepark} and \textit{airplane}, respectively.
    }
    \label{fig:ptb-multi}
\end{figure*}

\subsection{Implementation Details}
\subsubsection{AMN-S}
The maximal magnitude of perturbation $\epsilon$ under $l_p$-norm is set to be $\frac{16}{255}$ in all experiments for fair comparison.
AMN-S only perturbs one neuron at a time, which is randomly selected.
We use stochastic gradient descent (SGD) for perturbation optimization and employ a step size decay strategy, which starts with an initial step size of $4\epsilon$ and is halved every two epochs ($N_{drop}=2$). 
A mini-batch strategy is adopted for the generation set, where the batch size is set to 10, and the number of optimization epochs is fixed at 10.

The desired value of target neuron $t$ is set based on the statical characteristics of neuron values. 
Specifically, we calculate the mean ($\mu$) and standard deviation ($\sigma$) of different neurons on images from the ImageNet-1K validation set.
With preliminary experiments with ResNet50 and the specified target neuron, we find out that large $t$ doesn't ensure a strong attack, and the best attack appears when $t = \mu + 10\sigma$ as shown in Table~\ref{table:target_value}.

\subsubsection{ANM-M}
Basically, ANM-M is implemented in a similar way as ANM-S. 
The key difference is that the goal of ANM-M is to attack a set of neurons.
The number of neurons is set to be 12 in ANM-M, which has the best attack performance according to our testing on ResNet18 (see Fig.~\ref{fig:num_neuron}).
Considering that the value range of neurons are different, the target value $t$ in ANM-M are decided neuron by neuron and aligned with the findings in ANM-S.

\begin{figure}[t]
    \centering
    \begin{subfigure}[t]{0.24\linewidth}
        \centering
        \includegraphics[width=\linewidth]{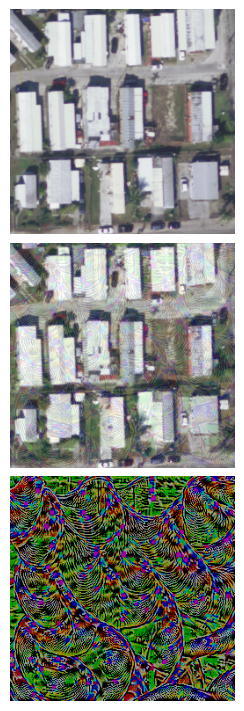}
        \caption{ResNet50}
        \label{fig:ptb-multi-ucm_a}
    \end{subfigure}
    \hspace*{-10pt}
    \begin{subfigure}[t]{0.24\linewidth}
        \centering
        \includegraphics[width=\linewidth]{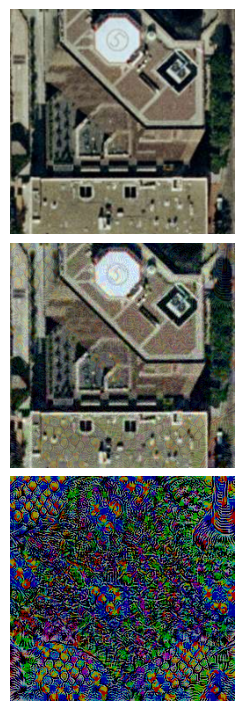}
        \caption{ResNet18}
        \label{fig:ptb-multi-ucm_b}
    \end{subfigure}
    \hspace*{-10pt}
    \begin{subfigure}[t]{0.24\linewidth}
        \centering
        \includegraphics[width=\linewidth]{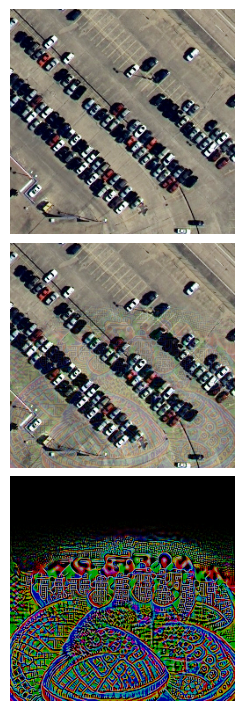}
        \caption{VGG19}
        \label{fig:ptb-multi-ucm_c}
    \end{subfigure}
    \hspace*{-10pt}
    \begin{subfigure}[t]{0.24\linewidth}
        \centering
        \includegraphics[width=\linewidth]{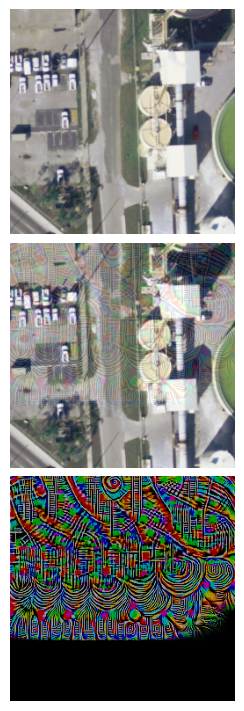}
        \caption{VGG11}
        \label{fig:ptb-multi-ucm_d}
    \end{subfigure}
    
    \caption{
       Examples of adversarial perturbations generated by ANM-M and applied to the UCM dataset. From top to bottom, there are original clean images, adversarial examples and adversarial perturbations generated on ImageNet (amplified 10x for visibility). }
    \label{fig:ptb-multi-ucm}
\end{figure}



\subsection{Experimental Results}\label{subsec:result}
The experimental results of ANM-S and ANM-M are summarized in Table~\ref{table:results}, where we evaluate four different models ResNet50, VGG19, ResNet18, and VGG11, against our ANM attack methods (ANM-S and ANM-M) across two datasets, UCM and AID. 
Note that the adversarial perturbations are generated without accessing downstream models and data. The only assumption is what pretrained models are used.
The effectiveness of each attack is measured by the classification accuracy (ACC) after perturbation, where a lower ACC indicates a more successful attack. The ACC under attack is averaged over multiple adversarial examples crafted using 10 pretrained perturbations and the testing datasets. This means that the number of adversarial examples tested for each UCM model is $1,050\times10=10,500$, and for AID models, the total is $5,000\times10=50,000$.
Additionally, the table highlights the percentage decrease in ACC relative to the baseline accuracy without any attack.

First, ANM-S induces substantial reductions in classification accuracy across all evaluated models, with average accuracy drops of 63.01\% on the UCM dataset and 69.74\% on the AID dataset across the four models. Examples of adversarial samples are illustrated in Fig.~\ref{fig:ptb-1-ucm}.
As previously discussed, ANM-S is designed to amplify targeted neuron values, leading to performance degradation in victim models. To demonstrate this effect, we visualize the activation values of neurons in Fig.~\ref{fig:neuron_value_vis}, where both the pretrained model ($f_p$) and the downstream model ($f_d$) are ResNet50. The adversarial perturbations are generated on $f_p$ and subsequently applied to $f_d$.
Fig.~\ref{fig:neuron_value_vis_a} displays the activation values of the 2015th neuron in ResNet50 ($f_p$) when processing ImageNet-1K. A clear distinction is evident between the neuron values for clean samples (blue) and adversarial examples (red) generated by ANM-S. Similar patterns are observed in Fig.~\ref{fig:neuron_value_vis_b} and Fig.~\ref{fig:neuron_value_vis_c}, where adversarial examples created by ANM-S targeting $f_p$ are used to attack $f_d$ on the UCM dataset. These results demonstrate that ANM-S effectively activates target neurons without requiring access to downstream models.
Furthermore, Fig.~\ref{fig:comp} compares the neuron values of target and non-target neurons under \mbox{ANM-S} attacks. It is evident that the values of target neurons change dramatically, whereas other neurons exhibit relatively minor variations. Additionally, Fig.~\ref{fig:anm-s-values} showcases samples of ANM-S targeting various neurons. ANM-S is capable of activating any targeted neuron, leading to recognition failures in victim models. This highlights the flexibility and robust attack capabilities of ANM-S.

Second, as expected, attacking multiple neurons enhances the attack capabilities of ANM-S. Compared to ANM-S, ANM-RANDOM results in an accuracy drop of up to 81.43\% on the UCM dataset and up to 80.98\% on the AID dataset across the four models. 
We, however, also notice that ANM-RANDOM do not consistently result in greater accuracy degradation than ANM-S (see the result of ResNet18 on AID dataset). 
As discussed in Section~\ref{subsec:ptbmulti}, not all neurons contribute equally to the representation, and randomly selected neurons may hinder the performance of our ANM attack. 
To this end, we propose ANM-M, in which the target neurons are carefully selected through MIMS(mentioned in Section~\ref{subsec:ptbmulti}), and strong neuronal correlations are achieved.
Therefore, it is unsurprising that ANM-M achieves the best attack capabilities among these methods, and the results strongly supported the efficacy of our methodology.
We also compare the targeted neuron values of the same models as with ANM-S (see Fig.~\ref{fig:ptb-multi}). It is evident that ANM-M significantly amplifies the targeted neuron values, resulting in successful attacks. Additionally, more samples of adversarial examples and perturbations generated by ANM-M can be found in Fig.~\ref{fig:ptb-multi-ucm}. These results demonstrate that ANM-M can effectively targets specific neurons, leading to substantial classification errors and highlighting the robustness and flexibility of our attack strategy.

\begin{table*}[t]
\caption{Cross-model Attack Performance (Classification Accuracy \%) of ANM. under Our Attacks. The absolute accuracy drop from clean accuracy is indicated in \textcolor{red}{red} (the higher, the better). The best attack is highlighted in \textbf{BOLD} (the lower, the better).}
\label{table:trans-anm}
\centering

\begin{tabular}{ccccccccccccc}
\noalign{\hrule height.9pt}
\multirow{2}*{Dataset} & \multirow{2}*{\backslashbox{\begin{math} f_d \end{math}}{\begin{math} f_p\end{math}}}& -&\multicolumn{2}{c}{ResNet18} & \multicolumn{2}{c}{ResNet50} & \multicolumn{2}{c}{VGG11} & \multicolumn{2}{c}{VGG19} \\

&&Clean(\%)& ANM-S & ANM-M & ANM-S & ANM-M &ANM-S & ANM-M &ANM-S & ANM-M \\
\noalign{\hrule height.7pt}
\\[-0.9em]
\multirow{13}*{UCM}
&ResNet18 &93.24 & - & - &\makecell{40.18 \\ \textcolor{red}{($\downarrow$53.06)}}& \makecell{\textbf{33.36} \\ \textcolor{red}{($\downarrow$59.88
)}}&\makecell{75.90  \\ \textcolor{red}{($\downarrow$17.34
)}}& \makecell{57.54 \\ \textcolor{red}{($\downarrow$35.70
)}}& \makecell{72.34  \\ \textcolor{red}{($\downarrow$20.90
)}}& \makecell{53.39 \\ \textcolor{red}{($\downarrow$39.85
)}}\\
\\[-0.9em]
\hhline{~----------}
\\[-0.9em]
& ResNet50 & 94.48& \makecell{45.68\\ \textcolor{red}{($\downarrow$48.80)}}& \makecell{\textbf{38.95} \\ \textcolor{red}{($\downarrow$55.53)}} &-& - &\makecell{81.21 \\ \textcolor{red}{($\downarrow$13.27)}}& \makecell{64.20 \\ \textcolor{red}{($\downarrow$30.28)}}&\makecell{78.24 \\ \textcolor{red}{($\downarrow$16.24)}}&  \makecell{61.00\\ \textcolor{red}{($\downarrow$33.48)}}\\
\\[-0.9em]
\hhline{~----------}
\\[-0.9em]
& VGG11&90.76 & \makecell{51.49 \\ \textcolor{red}{($\downarrow39.27$)}} & \makecell{37.61 \\ \textcolor{red}{($\downarrow$53.15)}} & \makecell{53.78\\ \textcolor{red}{($\downarrow$36.98)}} & \makecell{46.33\\ \textcolor{red}{($\downarrow$44.43)}}  & -& - &\makecell{54.50 \\ \textcolor{red}{($\downarrow$36.26)}} & \makecell{\textbf{33.39}\\ \textcolor{red}{($\downarrow$57.37)}} \\
\\[-0.9em]
\hhline{~----------}
\\[-0.9em]
& VGG19&91.33 & \makecell{66.69\\ \textcolor{red}{($\downarrow24.64$)}}& \makecell{65.67\\ \textcolor{red}{($\downarrow25.66$)}}& \makecell{66.36\\ \textcolor{red}{($\downarrow24.97$)}}&  \makecell{66.03\\ \textcolor{red}{($\downarrow25.30$)}}& \makecell{67.07\\ \textcolor{red}{($\downarrow24.26$)}}& \makecell{\textbf{34.69}\\ \textcolor{red}{($\downarrow56.64$)}}&-&-\\
\\[-0.9em]
\hhline{~----------}
\\[-0.9em]
& EfficientNet B0&90.29 & \makecell{56.85 \\ \textcolor{red}{($\downarrow33.44$)}}& \makecell{\textbf{41.42}\\ \textcolor{red}{($\downarrow48.87$)}}& \makecell{58.30 \\ \textcolor{red}{($\downarrow31.99$)}}& \makecell{44.66\\ \textcolor{red}{($\downarrow45.63$)}}& \makecell{78.97 \\ \textcolor{red}{($\downarrow11.32$)}}&\makecell{67.06\\ \textcolor{red}{($\downarrow23.23$)}}& \makecell{77.28\\ \textcolor{red}{($\downarrow13.01$)}}&\makecell{59.40 \\ \textcolor{red}{($\downarrow30.89$)}}\\
\\[-0.9em]
\hhline{~----------}
\\[-0.9em]
& MobileNet L &93.81 & \makecell{66.62 \\ \textcolor{red}{($\downarrow27.19$)}}&\makecell{\textbf{58.91}\\ \textcolor{red}{($\downarrow34.90$)}}& \makecell{72.37\\ \textcolor{red}{($\downarrow21.44$)}}& \makecell{62.15\\ \textcolor{red}{($\downarrow31.66$)}} &\makecell{86.10\\ \textcolor{red}{($\downarrow\ 7.71$)}}& \makecell{72.78\\ \textcolor{red}{($\downarrow21.03$)}} & \makecell{82.97 \\ \textcolor{red}{($\downarrow10.84$)}}& \makecell{59.40\\ \textcolor{red}{($\downarrow34.41$)}}\\
\\[-0.9em]
\hhline{~----------}
\\[-0.9em]
& Average Drop & -&\textcolor{red}{$\downarrow44.43$}& \textcolor{red}{$\downarrow51.89$}&\textcolor{red}{$\downarrow43.82$} &\textcolor{red}{$\downarrow50.23$} & \textcolor{red}{$\downarrow27.44$}&\textcolor{red}{$\downarrow42.94$}&\textcolor{red}{$\downarrow31.43$}&\textcolor{red}{$\downarrow47.89$} \\
\\[-0.9em]
\noalign{\hrule height.7pt}
\\[-0.9em]
\multirow{13}*{AID} 
&   ResNet18 &89.66 & -&-& \makecell{23.52\\ \textcolor{red}{($\downarrow66.14$)}}&  \makecell{\textbf{19.79}\\ \textcolor{red}{($\downarrow69.87$)}}& \makecell{66.87\\ \textcolor{red}{($\downarrow22.79$)}}& \makecell{ 42.48\\ \textcolor{red}{($\downarrow47.18$)}}&\makecell{62.05\\ \textcolor{red}{($\downarrow27.61$)}}& \makecell{ 38.02\\ \textcolor{red}{($\downarrow51.64$)}}\\
\\[-0.9em]
\hhline{~----------}
\\[-0.9em]
&ResNet50 &90.44 &\makecell{25.42\\ \textcolor{red}{($\downarrow65.02$)}}&\makecell{\textbf{19.95} \\ \textcolor{red}{($\downarrow70.49$)}}& -&-& \makecell{68.15\\ \textcolor{red}{($\downarrow22.29$)}}& \makecell{42.46\\ \textcolor{red}{($\downarrow47.98$)}}&\makecell{62.27\\ \textcolor{red}{($\downarrow28.17$)}}& \makecell{38.08\\ \textcolor{red}{($\downarrow52.36$)}}\\
\\[-0.9em]
\hhline{~----------}
\\[-0.9em]
&VGG11&89.80 & \makecell{28.42\\ \textcolor{red}{($\downarrow61.38$)}}&\makecell{19.50\\ \textcolor{red}{($\downarrow70.30$)}}& \makecell{31.17\\ \textcolor{red}{($\downarrow58.63$)}}&\makecell{23.09\\ \textcolor{red}{($\downarrow66.71$)}}& -&-&\makecell{39.92\\ \textcolor{red}{($\downarrow49.88$)}}&\makecell{\textbf{18.90}\\ \textcolor{red}{($\downarrow70.90$)}}\\
\\[-0.9em]
\hhline{~----------}
\\[-0.9em]
&VGG19&87.40 & \makecell{48.58\\ \textcolor{red}{($\downarrow38.82$)}}&\makecell{47.40\\ \textcolor{red}{($\downarrow40.00$)}}&\makecell{ 48.45\\ \textcolor{red}{($\downarrow38.95$)}}&  \makecell{47.65\\ \textcolor{red}{($\downarrow39.75$)}}& \makecell{49.13\\ \textcolor{red}{($\downarrow38.27$)}}& \makecell{\textbf{24.14}\\ \textcolor{red}{($\downarrow63.26$)}}&-&-\\
\\[-0.9em]
\hhline{~----------}
\\[-0.9em]
& EfficientNet B0&84.98 & \makecell{32.27\\ \textcolor{red}{($\downarrow52.71$)}}&\makecell{\textbf{21.20} \\ \textcolor{red}{($\downarrow63.78$)}}& \makecell{31.38\\ \textcolor{red}{($\downarrow53.60$)}}&\makecell{23.00 \\ \textcolor{red}{($\downarrow61.89$)}}& \makecell{71.10 \\ \textcolor{red}{($\downarrow13.88$)}}&\makecell{52.03\\ \textcolor{red}{($\downarrow32.95$)}}&  \makecell{67.53\\ \textcolor{red}{($\downarrow17.45$)}}&\makecell{44.63\\ \textcolor{red}{($\downarrow40.35$)}}\\
\\[-0.9em]
\hhline{~----------}
\\[-0.9em]
& MobileNet L&88.92 &  \makecell{45.06 \\ \textcolor{red}{($\downarrow43.86$)}}&\makecell{\textbf{36.62}\\ \textcolor{red}{($\downarrow52.30$)}}& \makecell{49.42 \\ \textcolor{red}{($\downarrow39.50$)}}&\makecell{39.50\\ \textcolor{red}{($\downarrow49.42$)}}& \makecell{67.07 \\ \textcolor{red}{($\downarrow21.85$)}}&\makecell{47.95\\ \textcolor{red}{($\downarrow40.97$)}}& \makecell{61.80\\ \textcolor{red}{($\downarrow27.12$)}}&\makecell{40.50\\ \textcolor{red}{($\downarrow48.42$)}} \\
\\[-0.9em]
\hhline{~----------}
\\[-0.9em]
& Average Drop & -&\textcolor{red}{$\downarrow58.58$}& \textcolor{red}{$\downarrow64.42$}&\textcolor{red}{$\downarrow57.88$} &\textcolor{red}{$\downarrow63.03$} & \textcolor{red}{$\downarrow34.81$}&\textcolor{red}{$\downarrow53.69$}&\textcolor{red}{$\downarrow39.61$}&\textcolor{red}{$\downarrow58.51$} \\
\\[-0.9em]
\noalign{\hrule height.9pt}
\end{tabular}
\end{table*}

\subsection{Transferability of ANM}
In the previous section, we explored the adversarial capabilities of ANM when the victim models share certain layers with the pretrained models. This scenario is common in transfer learning applications where models are fine-tuned from a shared backbone. However, in real-world applications, victim models may often be entirely different architectures or trained independently from the pretrained models used to generate adversarial examples. To evaluate the robustness and generalizability of ANM under such conditions, we conducted experiments where the victim models do not share any architectural or training similarities with the pretrained models. We additionally retrained the four downstream classifiers for testing, 2 MobileNet V3 Large \cite{howard2019searching} and 2 EfficientNet B0 \cite{tan2019efficientnet} for UCM and AID seperately. The results in Table \ref{table:trans-anm} are the overall accuracy over 10,500 adversarial examples for UCM models and 50,000 examples for AID models.

The results presented in Table \ref{table:trans-anm} clearly demonstrate that both ANM-S and ANM-M are highly transferable adversarial attacks with substantial attack capabilities across diverse neural network architectures.
On average, ANM-M induces a more significant accuracy drop of 64.42\% compared to ANM-S, which results in an average reduction of 58.58\%. This consistent trend across both the UCM and AID datasets underscores the robustness and generalizability of these attack methods. The high transferability indicates that adversarial examples generated by ANM-S and ANM-M are effective not only against the models they were specifically designed for but also against a wide range of other architectures, highlighting their potential as pervasive threats in various deployment scenarios.

A closer inspection of the table reveals that ANM-M consistently outperforms ANM-S across all evaluated models, affirming the advantage of manipulating multiple neurons simultaneously. 
For instance, on the AID dataset, VGG11 experiences a dramatic accuracy drop of 
70.30\% under ANM-M compared to 61.38\% under ANM-S. Similarly, ResNet50 on the AID dataset shows a substantial decrease of 70.49\% with ANM-M versus 65.02\% with ANM-S. These significant reductions highlight that targeting multiple neurons amplifies the attack's effectiveness, likely due to the compounded disruption of interconnected neural pathways. Conversely, models like EfficientNet B0 exhibit relatively lower vulnerability, with ANM-M causing a 63.78\% drop compared to 52.71\% with ANM-S on the AID dataset. This variation suggests that while ANM-S and ANM-M are broadly effective, the degree of susceptibility can vary based on the architectural intricacies of each model, with deeper and more complex architecture like ResNet50 being particularly at risk when multiple neurons are manipulated.

The attack effectiveness of ANM-S and ANM-M also exhibits notable differences between the UCM and AID datasets, with both attack variants causing greater accuracy drops on the AID dataset. Specifically, the average accuracy drop for ANM-M on AID is 64.42\%, compared to 51.89\% on UCM. This suggests that the AID dataset may possess inherent characteristics, such as higher intra-class variability or more complex feature distributions, that make models trained on it more susceptible to adversarial perturbations. For example, ResNet18 on AID experiences a 69.87\% accuracy drop under ANM-M, significantly higher than its counterpart on UCM (59.88\%). This dataset-specific vulnerability underscores the generalizability of ANM attacks, demonstrating their ability to effectively target models across different data distributions. Furthermore, the enhanced performance of ANM-M across both datasets highlights the critical role of multiple neuron manipulation in amplifying attack strategies, making ANM-M a more formidable adversarial tool compared to its single-neuron counterpart.

In summary, the cross-model attack performance of ANM underscores its potent ability to undermine the integrity of diverse neural network architectures across different datasets. The high transferability of ANM highlights the pressing need for robust defense mechanisms to safeguard machine learning systems against such sophisticated adversarial threats. 

\section{Conclusion}\label{sec:conclusion}

In this paper, we introduce a novel adversarial attack strategy tailored for transfer learning in remote sensing tasks, developing two distinct attack variants: ANM-S (Adversarial Neuron Manipulation for Single Neuron) and ANM-M (Adversarial Neuron Manipulation for Multiple Neurons). Through extensive evaluations, we demonstrate that both ANM-S and ANM-M possess high transferability and formidable attack capabilities across a range of neural network architectures and datasets. 
These results underscore the inherent risks associated with adapting pretrained models in remote sensing, emphasizing the urgent need for developing resilient defense mechanisms. As transfer learning continues to be a cornerstone in deploying advanced remote sensing solutions, ensuring the robustness of these models against sophisticated adversarial attacks like ANM-S and ANM-M becomes paramount. 
Potential approaches include adversarial training specifically designed for remote sensing data, the implementation of model ensemble techniques to dilute the impact of adversarial perturbations, and the exploration of advanced detection mechanisms that can identify and mitigate adversarial examples in real-time. Additionally, further investigation into the architectural features that confer resilience or susceptibility to adversarial manipulations can inform the design of more robust neural network models. By addressing these challenges, the remote sensing community can better safeguard pretrained models against sophisticated adversarial threats, ensuring the continued effectiveness and trustworthiness of remote sensing technologies.





\bibliographystyle{IEEEtran}

\bibliography{bibfile}


\end{document}